%% file: neurips_2025.tex
\documentclass{article}

\usepackage[preprint]{neurips_2025}
\usepackage{amssymb}
\usepackage[utf8]{inputenc}
\usepackage[T1]{fontenc}
\usepackage[hidelinks]{hyperref}
\usepackage{url}
\usepackage{booktabs}
\usepackage{amsfonts}
\usepackage{amsmath}

\newcommand{\github}{\raisebox{-1.5pt}{\includegraphics[height=1.05em]{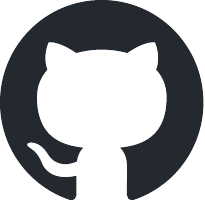}}}
\newcommand{\paperlogo}{\raisebox{-1.5pt}{\includegraphics[height=0.98em]{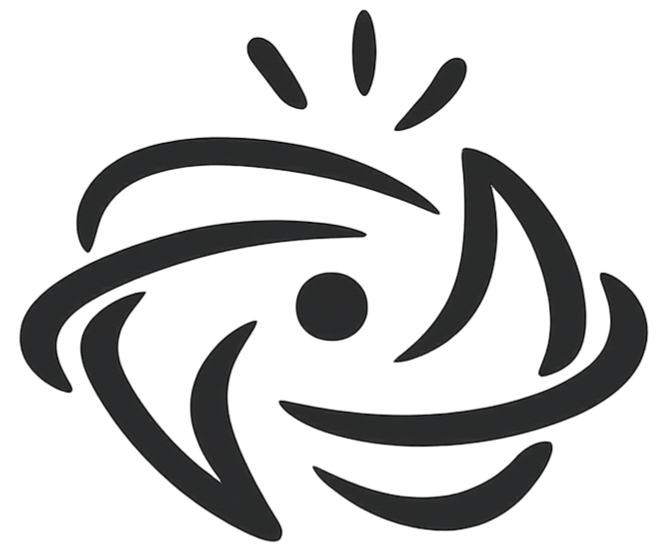}}}
\newcommand{\youtubeogo}{\raisebox{-1.5pt}{\includegraphics[height=0.98em]{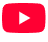}}}

\usepackage{nicefrac}
\usepackage{microtype}
\usepackage[table]{xcolor}
\usepackage{graphicx}
\usepackage{placeins} 
\usepackage{pdfpages}
\usepackage{capt-of} 

\title{\paperlogo \quad DeepReviewer 2.0:  A Traceable Agentic System for Auditable Scientific Peer Review}

\author{
\textbf{Yixuan Weng\textsuperscript{1}},
 \textbf{Minjun Zhu\textsuperscript{1,2}},
  \textbf{Qiujie Xie\textsuperscript{1,2}},
    \textbf{Zhiyuan Ning\textsuperscript{1}},
    \textbf{Shichen Li\textsuperscript{1,3}},
     \textbf{Panzhong Lu\textsuperscript{1}}, \\
 \textbf{Zhen Lin\textsuperscript{1}},
  \textbf{Enhao Gu\textsuperscript{1}},
 \textbf{Qiyao Sun\textsuperscript{1}},
 \textbf{Yue Zhang\textsuperscript{1}},
\\
\\
 \textsuperscript{1}Engineering School, Westlake University,
 \textsuperscript{2}Zhejiang University,
 \textsuperscript{3}Soochow University
\\
 \github{Code} : \href{https://github.com/ResearAI/DeepReviewer-v2}{ResearAI/DeepReviewer-v2} \quad\quad\quad\quad \youtubeogo \textbf{Video}:\href{https://www.youtube.com/watch?v=mMg5XzcaDCw}{Youtube:mMg5XzcaDCw} \\
  \paperlogo{} \textbf{Project}: \href{https://deepscientist.cc}{https://deepscientist.cc} \\ 
}

\begin{document}

\maketitle

\begin{abstract}
\input{sections/abstract}
\end{abstract}

\begin{center}
  \begin{minipage}{\textwidth}
    \centering
    \vspace{-0.35em}
    \includegraphics[width=0.99\textwidth]{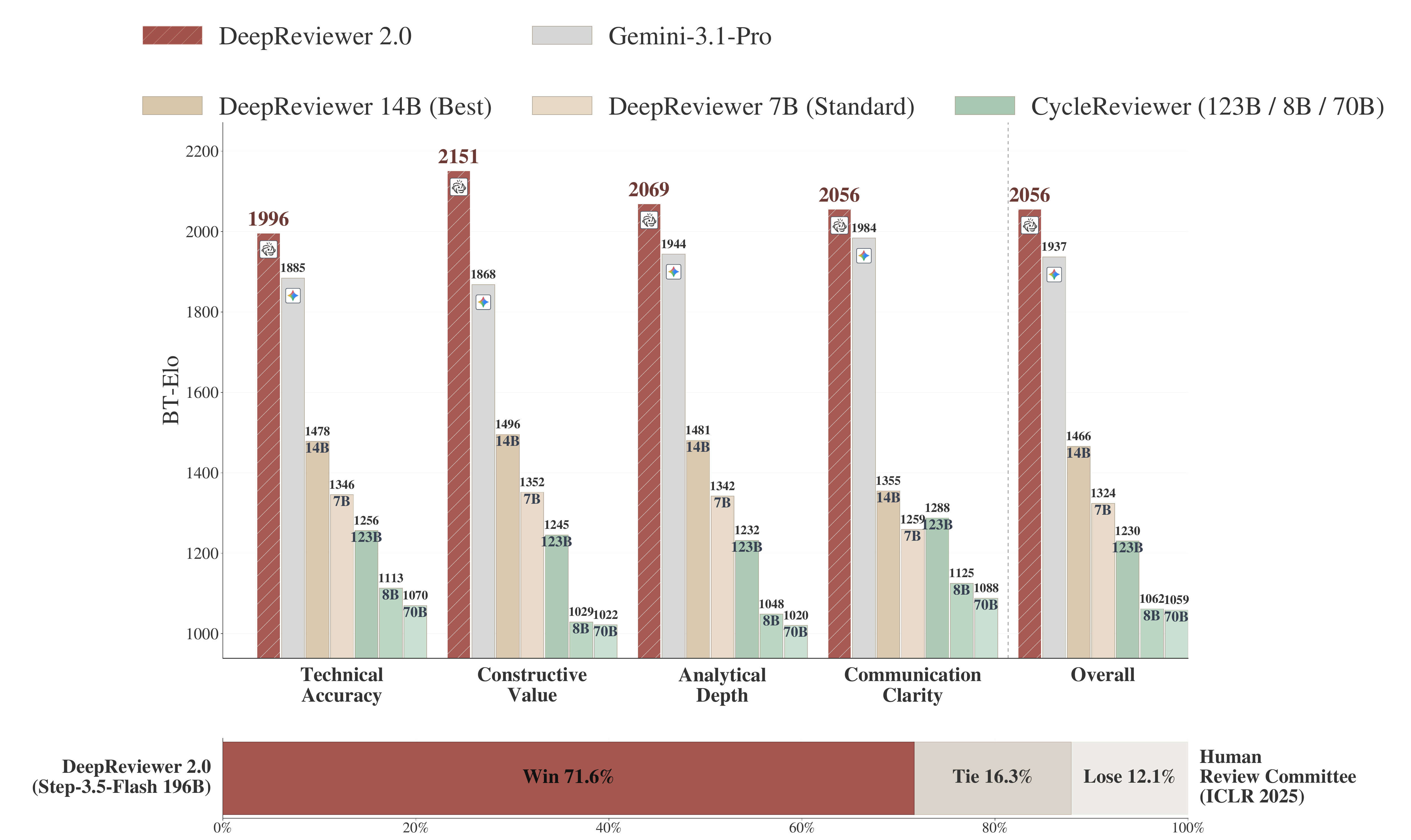}
    \vspace{-0.7em}
    \captionof{figure}{Dimension-wise Bradley--Terry Elo and blind preference against a human review committee.}
    \label{fig:dr2-vs-human}
  \end{minipage}
\end{center}
\vspace{-0.7em}

\section{Introduction}
\input{sections/introduction}

\section{Related work}
\input{sections/related_work}

\section{DeepReviewer 2.0}
\label{sec:method}
\input{sections/methodology}

\section{Experiments}
\input{sections/experiments}

\section{Conclusion}
\input{sections/conclusion}

\section{Limitations and Ethical Risk Considerations}
\label{sec:limitations}
\input{sections/limitations}

\bibliographystyle{iclr2026_conference}

{
  \catcode`\&=12\relax
  \bibliography{iclr2026_conference}
}

\clearpage
\appendix
\input{sections/appendix}

\end{document}

%% file: sections/abstract.tex
Automated peer review is often framed as generating fluent critique, yet reviewers and area chairs need judgments they can \emph{audit}: where a concern applies, what evidence supports it, and what concrete follow-up is required. DeepReviewer~2.0 is a process-controlled agentic review system built around an output contract: it produces a \textbf{traceable review package} with anchored annotations, localized evidence, and executable follow-up actions, and it exports only after meeting minimum traceability and coverage budgets. Concretely, it first builds a manuscript-only claim--evidence--risk ledger and verification agenda, then performs agenda-driven retrieval and writes anchored critiques under an export gate. On 134 ICLR~2025 submissions under three fixed protocols, an \emph{untuned 196B} model running DeepReviewer~2.0 outperforms Gemini-3.1-Pro-preview, improving strict major-issue coverage (37.26\% vs.\ 23.57\%) and winning 71.63\% of micro-averaged blind comparisons against a human review committee, while ranking first among automatic systems in our pool. We position DeepReviewer~2.0 as an assistive tool rather than a decision proxy, and note remaining gaps such as ethics-sensitive checks.

%% file: sections/introduction.tex
Peer review is not just a verdict; it is a justification that others must be able to trust. When a review claims ``the experiments are insufficient'' or ``there is novelty overlap,'' reviewers and area chairs still need to verify \emph{where} this applies in the manuscript and \emph{what} evidence supports the judgment before they can stand behind it. A usable review therefore has two properties: it is \emph{auditable} (concerns are anchored to concrete locations and evidence) and \emph{actionable} (it specifies concrete follow-up actions, such as missing baselines or required analyses). \textbf{Our work studies how to build automated reviewing around these properties rather than around fluent prose alone.}

Many automated reviewers are still built and evaluated mainly as one-shot generation or scoring systems. Even when the review sounds fluent, a common failure mode remains: the critique is hard to verify and hard to act on because it is not tied to the manuscript. If a system says ``the experiments are insufficient'' but cannot point to the relevant table, paragraph, or missing control, reviewers cannot safely endorse the judgment and authors do not know what to fix first. Recent LLM-based reviewers improve surface plausibility and guideline alignment, but most still do not \emph{require} explicit evidence links as part of the review output \citep{Yuan2021CanWA,Idahl2024OpenReviewerAS,Gao2025ReviewAgentsBT}.

DeepReviewer~2.0 targets a different output. Compared to DeepReviewer~1.0 \citep{Zhu2025DeepReviewIL}, which already explored agentic retrieval for novelty checking, DeepReviewer~2.0 makes traceability a first-class requirement rather than a best-effort behavior. \textbf{We operationalize peer review as an agentic cognitive chain that produces a traceable review package.} The package includes a structured report, page/paragraph-anchored annotations, a prioritized repair plan, and an explicit novelty/value assessment. In Stage~I, the system performs a manuscript-only pre-review and builds a claim--evidence--risk ledger. In Stage~II, it turns that ledger into a verification agenda, checks novelty against closely comparable prior work under a matched-setting gate, and writes anchored annotations with concrete repair actions. Final synthesis is gated: the system does not export a package until minimum process requirements for retrieval, structure, and anchored feedback are satisfied (Section~\ref{sec:method}; Table~\ref{tab:functional}).

This framing is intentionally conservative. \textbf{DeepReviewer~2.0 enforces process constraints, not semantic truth.} That distinction matters because automated reviewing can fail in ways that are not visible from prose quality alone, including hidden prompt injection in PDFs \citep{Ye2024AreWT}. It also matters at deployment time: pre-review tools can be miscalibrated or overused when treated as substitutes for peer review rather than assistive systems \citep{Akella2025PrereviewTP}. By making evidence links explicit and conditioning export on minimum traceability requirements, we aim to improve what can be audited and executed, even when judgments remain fallible.

This leads to the central question of this paper: \textbf{Does making traceability and process constraints part of the review output contract improve diagnostic coverage and review usefulness under a fixed rubric?} We evaluate DeepReviewer~2.0 with three protocols aligned with this process view: strict issue coverage, anonymous ranking among automatic systems, and anonymous blind comparison against a human review committee under a shared rubric. Across these views, the results are consistent: DeepReviewer~2.0 ranks first among automatic systems in our pool, improves strict major-issue coverage over a Gemini-3.1-Pro baseline, and is preferred in blind comparison against the human committee.

\paragraph{\textbf{Contributions.}}
\begin{itemize}
  \item \textbf{Output contract.} We reframe automated reviewing around a traceable review-package interface (structured report, anchored annotations, repair plan, and novelty/value assessment), with traceability treated as a required output.
  \item \textbf{Process-controlled workflow.} We design a staged cognitive chain with an explicit claim--evidence--risk ledger, matched-setting retrieval for novelty checks, and an export gate that enforces minimum traceability requirements.
  \item \textbf{Three-protocol evaluation.} We evaluate DeepReviewer~2.0 with strict issue coverage, anonymous system ranking with Bradley--Terry strength estimation, and a blind comparison against a human review committee under the same rubric; under this fixed-rubric setup, DeepReviewer~2.0 improves strict major-issue coverage over a Gemini-3.1-Pro baseline, ranks first among automatic systems in our pool, and is preferred in blind comparison against the human committee (Tables~\ref{tab:strict-coverage} and \ref{tab:dr2-vs-human}).
\end{itemize}
\noindent\textbf{Positioning.} DeepReviewer~2.0 is intended as an assistive tool for reviewers and authors, not a replacement for peer review.

%% file: sections/related_work.tex
Automated scientific reviewing predates modern large language models, from early proposals for semi-automated review infrastructure to neural review generation \citep{Alicea2013ASP,Yuan2021CanWA}. Recent LLM-based systems improve review fluency and often add rubrics, role decomposition, or stronger training/evaluation resources, as in OpenReviewer and ReviewAgents \citep{Idahl2024OpenReviewerAS,Gao2025ReviewAgentsBT}. Adjacent work also studies aggregation/meta-reviewing and audits failure modes of LLM-written reviews under realistic conditions \citep{Hasan2024DeepTL,Li2025UnveilingTM}. \textbf{Unlike this line of work, DeepReviewer~2.0 is not centered on fluent review generation alone; it targets a traceable review package as the primary output.}

\begin{table*}[t]
\centering

\resizebox{\textwidth}{!}{%
\begin{tabular}{lccccccc}
\toprule
System & Full review & Retrieval & Novelty check & Claim check & Page/para notes & Structured package & Traceable evidence \\
\midrule
\citet{Yuan2021CanWA} & \checkmark & $\times$ & $\times$ & $\times$ & $\times$ & $\circ$ & $\times$ \\
OpenReviewer \citep{Idahl2024OpenReviewerAS} & \checkmark & $\times$ & $\times$ & $\times$ & $\times$ & \checkmark & $\times$ \\
ReviewAgents \citep{Gao2025ReviewAgentsBT} & \checkmark & $\circ$ & $\circ$ & $\times$ & $\times$ & \checkmark & $\circ$ \\
AgentReview \citep{Jin2024AgentReviewEP} & $\circ$ & $\times$ & $\times$ & $\times$ & $\times$ & $\circ$ & $\times$ \\
PaperReview.ai \citep{jiangng2025techoverview} & \checkmark & \checkmark & \checkmark & $\circ$ & $\times$ & \checkmark & $\times$ \\
DeepReviewer 1.0 \citep{Zhu2025DeepReviewIL} & \checkmark & \checkmark & \checkmark & $\circ$ & $\times$ & $\circ$ & $\circ$ \\
\textbf{DeepReviewer 2.0 (ours)}  & \checkmark & \checkmark & \checkmark & \checkmark & \checkmark & \checkmark & \checkmark \\
\bottomrule
\end{tabular}
}
\caption{Functional comparison of representative automated reviewing systems (yes/partial/no). Here \checkmark\ denotes an explicit, first-class capability in the system interface or output contract; $\circ$ denotes partial/limited support; and $\times$ denotes absence. Assignments follow capabilities explicitly described by the original interfaces/outputs, not inferred latent behavior; we mark \checkmark\ only when a capability is required by the interface/output contract, not merely achievable via free-form prompting. A fluent ``full review'' is not the same output target as an evidence-bound review package with traceable anchors.}
\label{tab:functional}
\end{table*}

A second thread treats reviewing as a process rather than a single generated artifact. AgentReview simulates reviewer--author--chair interactions to study how bias, expertise, and social dynamics affect outcomes \citep{Jin2024AgentReviewEP}. DeepReview decomposes reviewing into stages, including retrieval and verification for novelty checking \citep{Zhu2025DeepReviewIL}, while ReviewEval argues for multidimensional assessment of review quality (e.g., factuality and constructiveness) instead of a single score \citep{Kirtani2025ReviewEvalAE}. \textbf{DeepReviewer~2.0 follows this process view, but makes anchored annotations and export-time traceability constraints mandatory parts of the system contract.}

Our work is also informed by evidence-grounded scholarly assistance, novelty assessment, and document grounding. Retrieval-augmented scholarly assistants and citation-quality studies show that usefulness depends on local evidence support \citep{wengcycleresearcher, weng2026deepscientist}, not only document-level relevance \citep{Asai2024OpenScholarSS,Roy2024ILCiteREI}. Novelty assessment has been operationalized as contribution-level comparison pipelines with explicit taxonomies \citep{Wang2023SciMONSI,Zhao2025ARO}, and paragraph/page-level localization relies on layout-aware document modeling and datasets \citep{Xu2019LayoutLMPO,Pfitzmann2022DocLayNetAL}. \textbf{DeepReviewer~2.0 combines these ingredients for a different goal: an auditable peer-review artifact that jointly supports diagnosis, novelty checking, and actionable revision guidance.}

%% file: sections/methodology.tex
DeepReviewer 2.0 is built around a simple thesis: a usable peer review is not a monologue, but a sequence of constrained cognitive moves. A move may extract the paper's core claims, check novelty against closely comparable prior work, or write a localized critique with an executable repair action. These moves are grounded by concrete requirements: judgments must be traceable to anchors in the paper, comparisons must use matched settings, and critiques must end with actionable revisions. The system therefore produces an auditable review package rather than a single free-form review text.

At a high level, DeepReviewer 2.0 parses the PDF into anchored semantic units, performs an independent pre-review to build a claim--evidence--risk ledger and an investigation agenda, then runs agenda-driven verification while writing anchored annotations. Final synthesis is gated before exporting the review package. Figure~\ref{fig:method-overview} summarizes the workflow.

Implementation details (MCP tool interfaces, the modified OpenAI Agent SDK agent loop, and PASA integration) are deferred to Appendix~\ref{app:pdf-to-text} and Appendix~\ref{app:pseudocode}.

Stepping back, we define the task by three conditions that must hold together. \textbf{Soundness} asks whether major judgments close an internal evidence loop: each key critique must link to in-paper anchors (text spans, figures, tables, or equations), and missing anchors are treated as high-risk gaps. \textbf{Novelty/value} asks whether claimed contributions survive comparison to directly comparable prior work under matched settings; we record conservative novelty tags and default to ``unclear'' when evidence is insufficient. \textbf{Actionability} asks whether an author can act immediately: every anchored note ends with a concrete repair requirement (rewrite, clarification, ablation, added baseline, or reporting fix), not just a stylistic preference. DeepReviewer 2.0 enforces these as process constraints (schema, budgets, and traceability), not as semantic-truth guarantees.

\begin{center}
  \IfFileExists{figures/final.pdf}{%
    \includegraphics[width=\linewidth]{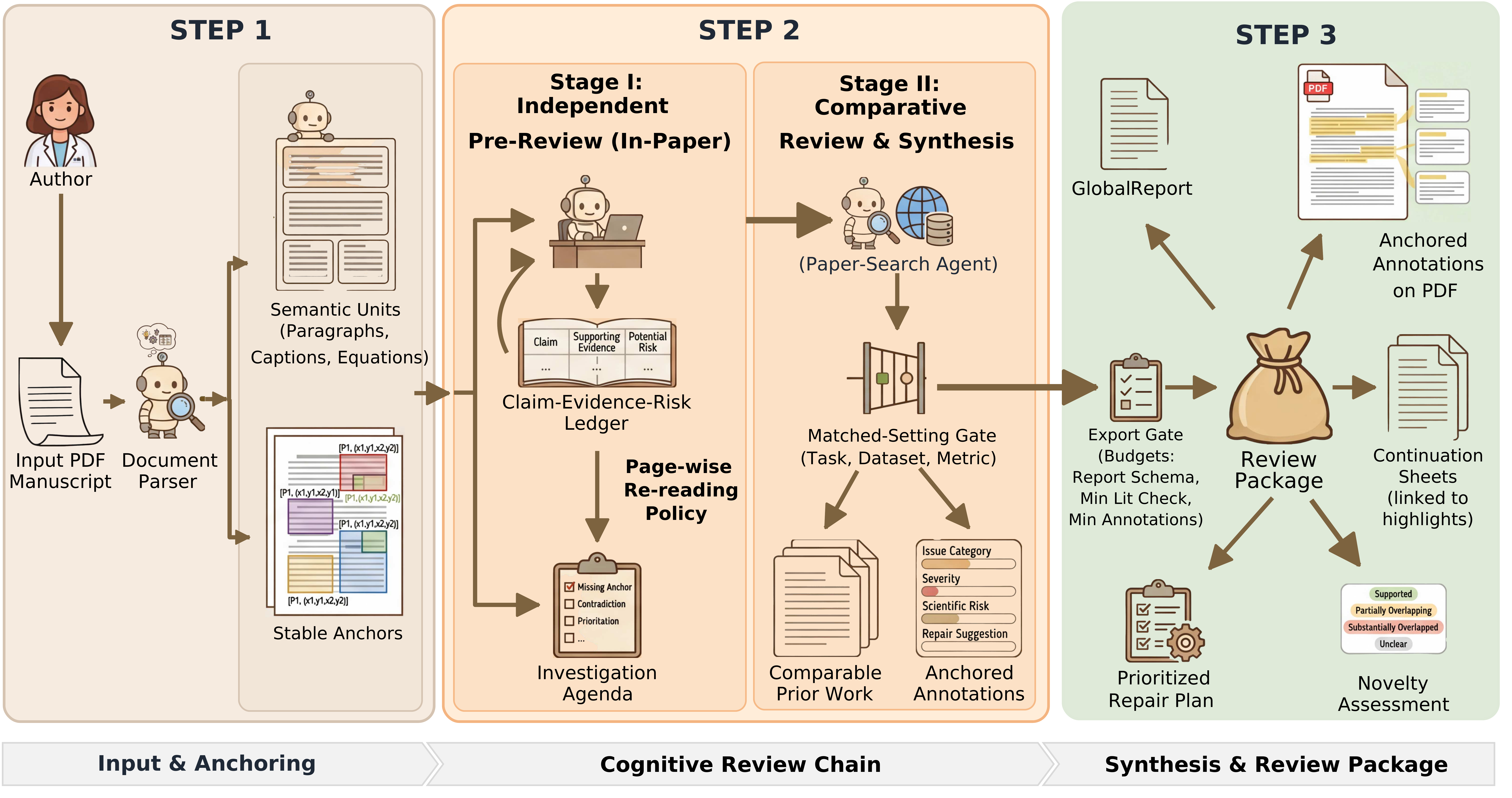}%
  }{%
    \fbox{%
      \parbox[c][0.18\textheight][c]{0.95\linewidth}{%
        \centering
        \textbf{Missing method overview figure.}\\
        Expected \texttt{figures/final.pdf}.
      }%
    }%
  }%
  \captionof{figure}{Overview of DeepReviewer~2.0 as an agentic cognitive chain. Stage~I parses the PDF into anchored units and builds a claim--evidence--risk ledger plus an investigation agenda. Stage~II performs agenda-driven verification (including matched-setting literature comparison) and writes anchored annotations with severity and concrete repair actions. Final synthesis is gated before exporting the review package.}
  \label{fig:method-overview}
\end{center}

\subsection{Task formulation and notation}

DeepReviewer 2.0 takes a paper in PDF form and turns it into a semantic representation that can be reasoned over while remaining traceable to the original layout. Concretely, we represent the paper as a sequence of semantic units (paragraphs, captions, equations with local context) paired with stable anchors back to the source. This view builds on progress in document layout understanding and scientific parsing \citep{Xu2019LayoutLMPO,Pfitzmann2022DocLayNetAL,Jimeno-Yepes2021ICDAR2C}; our contribution is to elevate anchors into a \textbf{required} part of the reviewing interface and downstream audit trail. We write the parsed paper as
\begin{equation}
  X=\{(u_i, a_i)\}_{i=1}^{N}, \qquad a_i=(p_i, b_i), \quad b_i \in \mathbb{R}_{+}^{4},
  \label{eq:paper-repr}
\end{equation}
where $u_i$ is the $i$-th semantic unit and $a_i$ stores its anchor: the page index $p_i$ and a bounding box $b_i=[x_{\min},y_{\min},x_{\max},y_{\max}]$ under the parser's page-local coordinate system.
In our implementation, anchors are stored primarily as a page-and-line-span identifier (page index plus a contiguous line/paragraph range), optionally augmented with bounding boxes when layout coordinates are available for rendering. Anchors are defined with respect to the parser output rather than PDF viewer rendering, which enables paragraph-level feedback to be traced back to concrete regions of the source PDF. Appendix~\ref{app:anchored-annotation} details the anchored representation and overlay rendering.

The output is not a single free-form review paragraph. DeepReviewer 2.0 outputs a review package that bundles global judgment, local evidence, and actionable next steps:
\begin{equation}
  Y=(\mathcal{R},\mathcal{A},\mathcal{P},\mathcal{N}).
  \label{eq:review-package}
\end{equation}
$\mathcal{R}$ is a structured final report (summary, strengths, weaknesses, overall judgment); $\mathcal{A}$ is a set of anchored annotations tied to specific paper locations; $\mathcal{P}$ is a prioritized repair plan; and $\mathcal{N}$ is a novelty/value assessment that makes comparisons explicit and records uncertainty when the evidence is thin. This output contract separates global synthesis from localized evidence, so readers can audit judgments without reconstructing the evidence trail manually.

\paragraph{\textbf{Explicit evidence trail.}}
DeepReviewer 2.0 treats the evidence trail as a first-class artifact, not as a latent reasoning trace. During the review, the system records typed links between (i) claims in the ledger, (ii) anchors in the paper, (iii) anchored annotations, and (iv) retrieved comparators used for novelty/value checks. We summarize this trail as a typed graph
\begin{equation}
  G=(V,E),\qquad
  V = V_{\text{claim}} \cup V_{\text{anchor}} \cup V_{\text{ann}} \cup V_{\text{prior}},
  \label{eq:evidence-graph}
\end{equation}
where edges encode relations such as \texttt{supported-by}, \texttt{contradicted-by}, \texttt{localized-to}, and \texttt{overlaps-with}. A minimal traceability constraint is that every anchored annotation links to at least one paper anchor in $V_{\text{anchor}}$, so that a reader can jump from critique to evidence without reconstructing the chain. In practice, these links also support traceability checks before export (e.g., validating that major notes and judgments have at least one concrete anchor).

\subsection{Stage I: Global diagnostic review}

Stage I is an independent pre-review. DeepReviewer 2.0 first builds a paper map without leaning on external literature or other reviewers' opinions. The output of this stage is a claim--evidence--risk ledger that makes the system commit to what it thinks the paper is claiming and what in-paper evidence is supposed to support those claims. Formally,
\begin{equation}
  \mathcal{L}^{(0)} = f_{\text{meta}}(X)
  = \{(c_k, E_k^{(0)}, r_k^{(0)}, s_k)\}_{k=1}^{K}, \qquad
  \mathcal{Q}=g(\mathcal{L}^{(0)}).
  \label{eq:ledger-agenda}
\end{equation}
Each ledger entry contains a core claim $c_k$, its initial in-paper evidence set $E_k^{(0)}$ (anchors into $X$), an initial risk statement $r_k^{(0)}$ (what would break scientifically if the evidence is missing or misinterpreted), and a status label $s_k$. At this stage we use $s_k \in \{\texttt{confirmed},\texttt{suspected}\}$ to separate defects that can be established from the manuscript alone versus issues that require verification. The stage also produces an investigation agenda $\mathcal{Q}$: a concrete list of questions to resolve in Stage II. This commits the system to explicit verification targets before final synthesis.

We keep the status criterion conservative. A defect is marked as ``confirmed'' when the paper provides a direct internal contradiction across anchors (e.g., inconsistent definitions or mismatched claims and tables), or when a core claim lacks any supporting anchor after a second pass. Otherwise it remains ``suspected'' and is elevated into the agenda. An example agenda item is: ``Which closely comparable papers already report the same evaluation setting as the main claim, and does the contribution still hold under that comparison?'' We prioritize agenda items by scientific risk: issues that can flip the validity of a central claim come first.

\paragraph{\textbf{Page-wise re-reading policy.}}
DeepReviewer 2.0 is not a single-pass generator. To reduce omission and to mitigate lost-in-the-middle failures, it iterates between selecting a paper span to (re)read and updating the ledger and annotations with the evidence found in that span:
\begin{equation}
  \pi_t = \mathrm{SelectSpan}(X, \mathcal{L}^{(t-1)}, \mathcal{Q}), \qquad
  (\mathcal{L}^{(t)}, \mathcal{A}^{(t)}) = \mathrm{Update}(\mathcal{L}^{(t-1)}, \mathcal{A}^{(t-1)}, X_{\pi_t}),
  \label{eq:reread-loop}
\end{equation}
where $\pi_t$ is a page-indexed location (page plus line/paragraph span). $\mathrm{SelectSpan}(\cdot)$ prioritizes spans implicated by high-risk ledger items and agenda questions, while still advancing in page order to maintain coverage. This is the core operational bias: keep synthesis tethered to localized evidence, and revisit uncertain spans before finalizing the verdict.
Here $\mathcal{A}^{(t)}$ denotes provisional anchored notes; final annotation writing and severity assignment are completed in Stage~II.

\subsection{Stage II: Verification-oriented annotation and synthesis}

Stage II turns the agenda into evidence and action. For each agenda item $q_m \in \mathcal{Q}$ ($m=1,\ldots,M$), DeepReviewer 2.0 retrieves and reads a candidate set of papers and uses them to test novelty/value claims and to sanity-check methodological choices. We write this as
\begin{equation}
  \mathcal{E}_m = \mathrm{Retrieve}(q_m), \qquad
  \hat{q}_m = \mathrm{Verify}(q_m, X, \mathcal{E}_m).
  \label{eq:retrieve-verify}
\end{equation}
In our implementation, $\mathrm{Retrieve}(\cdot)$ is carried out by an academic paper-search agent (PASA) that expands an agenda question into targeted queries and consolidates evidence across sources \citep{He2025PaSaAL}. We define $M=|\mathcal{Q}|$.

\paragraph{\textbf{Matched-setting comparability gate.}}
We apply a matched-setting comparability gate before using prior work as novelty-overlap evidence. A paper is treated as directly comparable only if it matches the same task definition, the same dataset/benchmark, and the same primary evaluation metric. Other factors (e.g., model scale or compute budget) are used only as secondary ordering cues when such metadata is available, but they are not part of the gate. When comparability fails (e.g., task or metric mismatch), we may still record retrieved papers as background context, but we do not treat them as overlap evidence against a claim.

\paragraph{\textbf{Conservative novelty tags.}}
Given a comparable set, $\mathrm{Verify}(\cdot)$ assigns one of four conservative novelty tags to each checked claim: \textsc{supported} (no directly comparable prior work covers the core mechanism/claim under the matched setting), \textsc{partially overlapping} (overlap exists but a clear residual novelty axis remains), \textsc{substantially overlapped} (directly comparable prior work already covers the core mechanism/claim under the matched setting), or \textsc{unclear}. The default is \textsc{unclear} when a claim cannot be mapped to a checkable mechanism or when evidence is insufficient or contradictory.

The goal is not generic similarity search, but matched-setting comparison under explicit comparability rules, aligned with evidence-grounded scholarly assistants and scientific fact-checking work \citep{Asai2024OpenScholarSS,Wang2023SciMONSI,Vladika2023ScientificFA}. DeepReviewer 2.0 uses this stage to update the ledger and to decide what can be stated confidently versus what should be flagged for follow-up experiments or clearer positioning.

The main output artifact of Stage II is a set of anchored annotations. Each annotation is written as a small, structured unit so that it can be audited and acted on. We model an annotation as
\begin{equation}
  a_j = (\ell_j, \tau_j, \sigma_j, \rho_j, \delta_j), \qquad a_j \in \mathcal{A}.
  \label{eq:annotation-unit}
\end{equation}
$\ell_j$ is the location anchor (page/paragraph, potentially with a bounding box); $\tau_j$ is the issue category; $\sigma_j$ is the severity (e.g., major/minor); $\rho_j$ explains the scientific risk (why it matters, not just that it is ``unclear''); and $\delta_j$ is the repair suggestion written so that an author can turn it into an explicit to-do. This bias toward actionability is consistent with annotation taxonomies and structured feedback settings that treat ``what should I do next?'' as the core value of critique \citep{Choudhary2022ReActAR,Tang2024SPHERESP}.
Operationally, we store $\ell_j$ as $(p_j,k_j^{\text{start}},k_j^{\text{end}})$ (page plus line span) and only then derive a highlight rectangle for rendering. This ordering matters: page-and-span anchors remain stable under small extraction variations (e.g., missing boxes or minor line segmentation differences) produced by the PDF-to-markdown parser, while the layout rectangle is best-effort and can fall back to a conservative approximation when bounding boxes are unavailable (Appendix~\ref{app:anchored-annotation}).

Finally, anchored notes are exported as a readable, human-facing artifact by rendering the anchors back onto the source document. We overlay translucent highlights over the anchored spans and add right-margin callouts containing diagnosis and repair actions; when margin space is insufficient, the full text is moved to a continuation sheet linked to the source highlight. Appendix~\ref{app:overlay} details the overlay rendering procedure.

\subsection{Process constraints and auditability guarantees}

DeepReviewer 2.0 is explicit about what it enforces and what it does not. The system is designed to enforce \textbf{process} properties: a structured report, minimum annotation coverage, and traceability links that let a reader check where a judgment came from. We can summarize the export gate for a review package as
\begin{equation}
  \begin{aligned}
    Y &= h\!\left(X, \mathcal{L}^{(*)}, \{\hat{q}_m\}_{m=1}^{M}, \mathcal{A}\right), \\
    \mathbf{1}_{\text{ready}}(Y) &= \mathbf{1}\!\left[\mathrm{Schema}(\mathcal{R})=1\right]
    \cdot \mathbf{1}\!\left[n_{\text{search}}\ge \alpha\right]
    \cdot \mathbf{1}\!\left[n_{\text{intent}}\ge \beta\right]
    \cdot \mathbf{1}\!\left[|\mathcal{A}|\ge \gamma\right].
  \end{aligned}
  \label{eq:synthesis-gate}
\end{equation}
Here $\mathcal{L}^{(*)}$ denotes the ledger after Stage~I/II updates, and $\mathrm{Schema}(\mathcal{R})$ checks that the final report follows the required structure (e.g., summary, strengths, weaknesses, prioritized issues, and actionable suggestions). $n_{\text{search}}$ and $n_{\text{intent}}$ enforce a minimum literature-checking budget and distinct question coverage for Stage~II; and $|\mathcal{A}|$ enforces that the system produced a minimum amount of anchored, actionable feedback. In implementation, readiness checks also validate anchor integrity and per-annotation field completeness (risk explanation and repair action), not only aggregate counts. In the runs reported in this paper, we use $\alpha=3$, $\beta=3$, and $\gamma=10$ as the enforced minimum budgets. In short, the goal is to make the output easier to audit, even if the underlying semantic judgments remain fallible.

The boundary is just as important. DeepReviewer 2.0 does not guarantee semantic truth, and it does not guarantee perfect novelty judgments. Automated reviewing is exposed to manipulation and systematic failure modes of language models, including prompt injection through hidden content in PDFs \citep{Ye2024AreWT}. It is also exposed to broader misalignment issues when a system designed for pre-review screening is treated as a substitute for peer review \citep{Akella2025PrereviewTP}. The right way to read DeepReviewer 2.0, therefore, is as a process-controlled agentic review system: it can help reviewers and authors by making judgments more traceable and feedback more executable, but it should not be treated as an oracle.

%% file: sections/experiments.tex
\subsection{Data, protocol, and metrics}

We evaluate on a sample of 134 ICLR~2025 submissions with three protocols aligned with our review-package contract: (i) strict issue coverage (diagnostic recall), (ii) anonymous chain ranking among automatic systems (end-to-end preference), and (iii) anonymous blind comparison against a human review committee under the same rubric. Strict coverage is computed on all 134 papers. Ranking metrics are computed on the subset where the Step-3 judge emits a valid comparison chain, and we use the same subset for both ranking protocols to avoid denominator drift (Appendix~\ref{app:notation}).

Before ranking, we apply fixed pipeline validity checks (e.g., malformed review-package exports or unparseable judge chains are excluded). This keeps the protocol executable but can understate end-to-end brittleness in deployment (Section~\ref{sec:limitations}).

Strict issue coverage follows a conservative two-step rubric: we extract a canonical issue list from the human committee packet with semantic deduplication, severity (major/minor), and a single category label (Step-1), then score each system by strict binary coverage of each issue (Step-2), counting missing outputs as uncovered.

For anonymous ranking (Step-3), a fixed judge (\texttt{gemini-3.1-pro-preview}) produces comparison chains with ties on five dimensions \citep{Kirtani2025ReviewEvalAE,Chu2024PREAP,Zhang2025UPMEAU}. We compute pairwise non-tie win fractions, uniform average win rate, tie-aware average rank, and Bradley--Terry Elo with 95\% paper-level bootstrap confidence intervals (1000 resamples) \citep{Selby2024PageRankAT}. The human baseline is a committee-style packet (all reviews and discussion + rebuttal + meta-review/decision).

\subsection{Results}

\noindent We organize the analysis by four research questions, but the evidence should be read jointly. RQ1 measures strict diagnostic coverage against human-identified issues. RQ2 tests end-to-end preference among automatic systems under a fixed rubric. RQ3 places the same review package under blind comparison with a human review committee. RQ4 explains where gains come from (and where they do not) via category-level coverage.

\paragraph{\textbf{RQ1: Does DeepReviewer 2.0 better cover human-identified issues under strict coverage, especially major issues?}}
Yes. Table~\ref{tab:strict-coverage} reports strict issue coverage against the Step-1 human issue list. Under this conservative benchmark, the most important signal is major-issue coverage: DeepReviewer~2.0 achieves the best strict overall coverage (35.74\%) and the best major-issue coverage (37.26\%), improving by 13.69 points over a Gemini-3.1-Pro baseline (23.57\%) and reducing the critical miss rate to 62.74\%. The larger gains on major issues match the system design: the claim--evidence--risk ledger, risk-prioritized agenda, and page-wise re-reading bias the process toward decision-changing defects rather than maximizing the count of minor comments. At the same time, the absolute miss rate remains high under a strict binary matching policy, so we interpret RQ1 as improved diagnostic recall under a conservative protocol, not as evidence of human-level completeness.
\input{tables/table_issue_coverage}
\FloatBarrier

\paragraph{\textbf{RQ2: Under anonymous ranking among automatic systems, is DeepReviewer 2.0 consistently preferred?}}
\noindent Yes. We summarize anonymous ranking among automatic systems with three complementary views (pairwise breadth, aggregate robustness, and metric agreement), each computed from the same Step-3 judge chains.

\begin{figure}[!htbp]
  \centering
  \includegraphics[width=\textwidth]{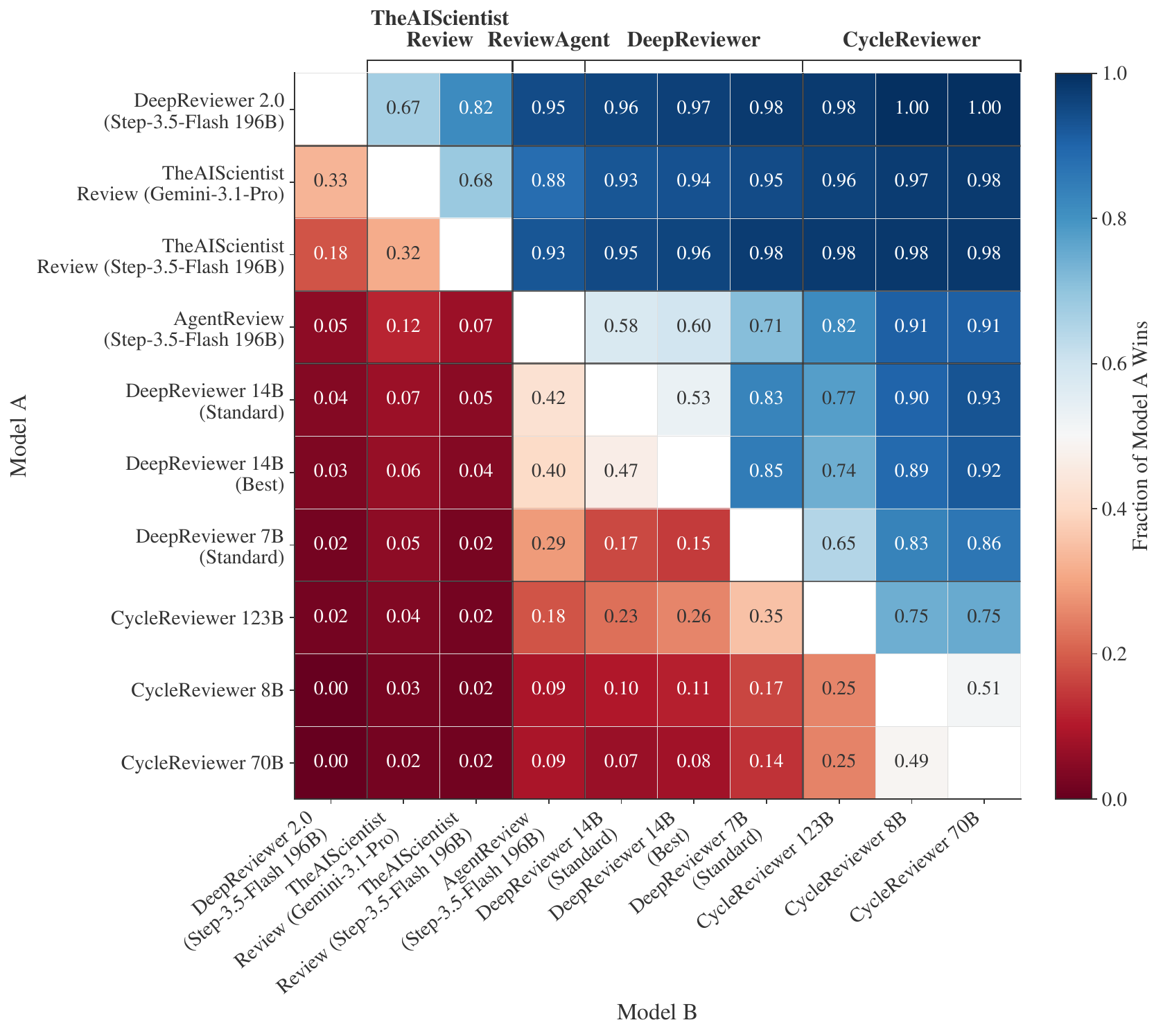}
  \caption{Pairwise non-tie win-fraction heatmap among automatic systems.}
  \label{fig:pairwise-heatmap}
\end{figure}
\FloatBarrier
\textbf{Pairwise breadth (Figure~\ref{fig:pairwise-heatmap}).} DeepReviewer~2.0 wins broadly across opponents; importantly, its hardest opponent in this pool is the Gemini-3.1-Pro baseline, against which it still wins 66.86\% of non-tied comparisons. This argues against a narrow gain that only holds against weak baselines.

\begin{figure}[t]
  \centering
  \includegraphics[width=0.98\textwidth]{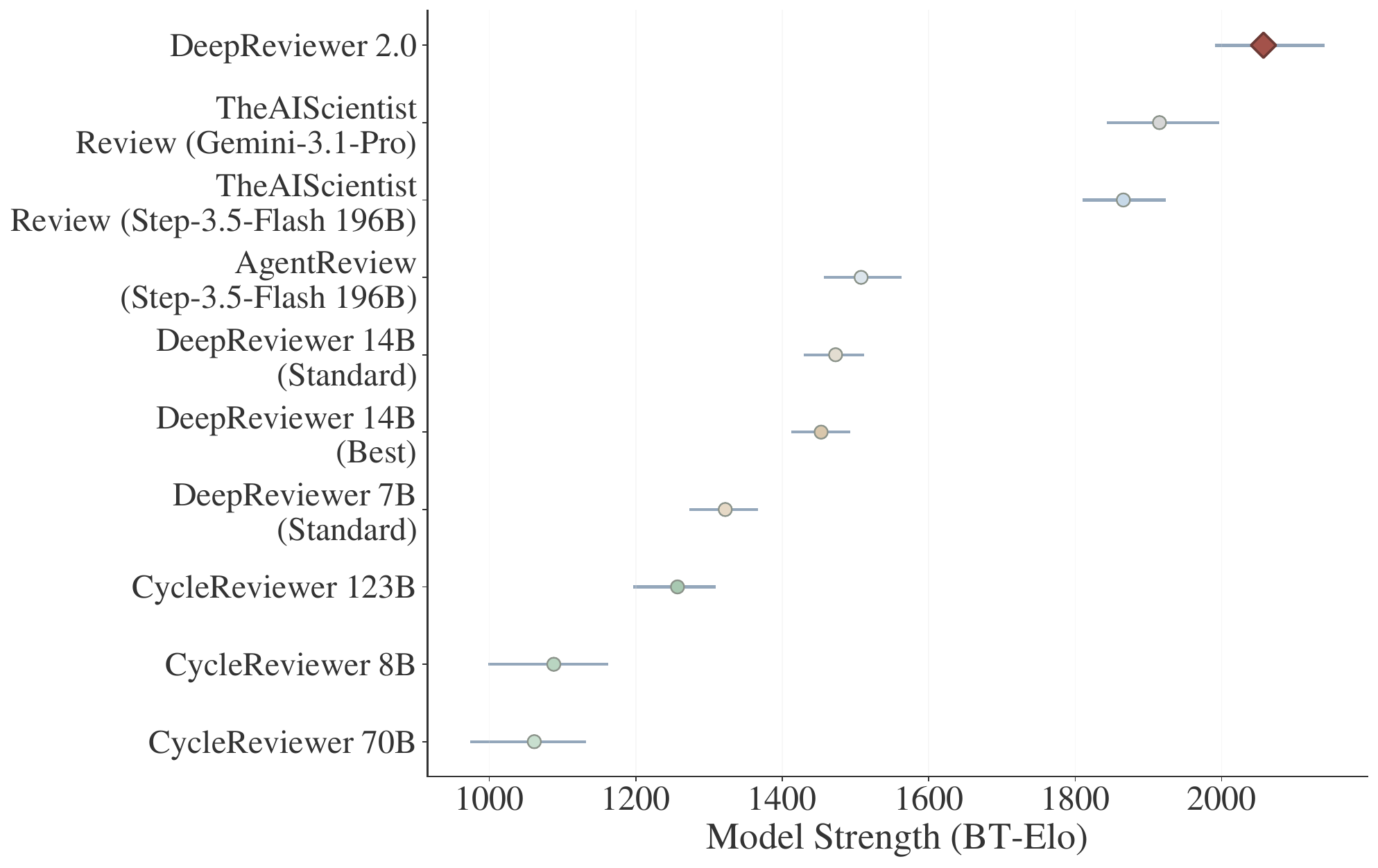}
  \caption{Bradley--Terry model strength estimation (Elo) with 95\% bootstrap confidence intervals (1000 paper-level resamples).}
  \label{fig:bt-elo}
\end{figure}

\textbf{Aggregate robustness (Figure~\ref{fig:bt-elo}).} Under an order-invariant strength model, DeepReviewer~2.0 remains the top-ranked system with 2057.17 Elo (95\% CI: [1990.55, 2140.81]), ahead of the strongest Gemini-3.1-Pro baseline (1915.11 Elo), with a 142.07 Elo gap. This supports the claim that preference is stable under aggregation and not an artifact of chain ordering.

\begin{figure}[t]
  \centering
  \includegraphics[width=\textwidth]{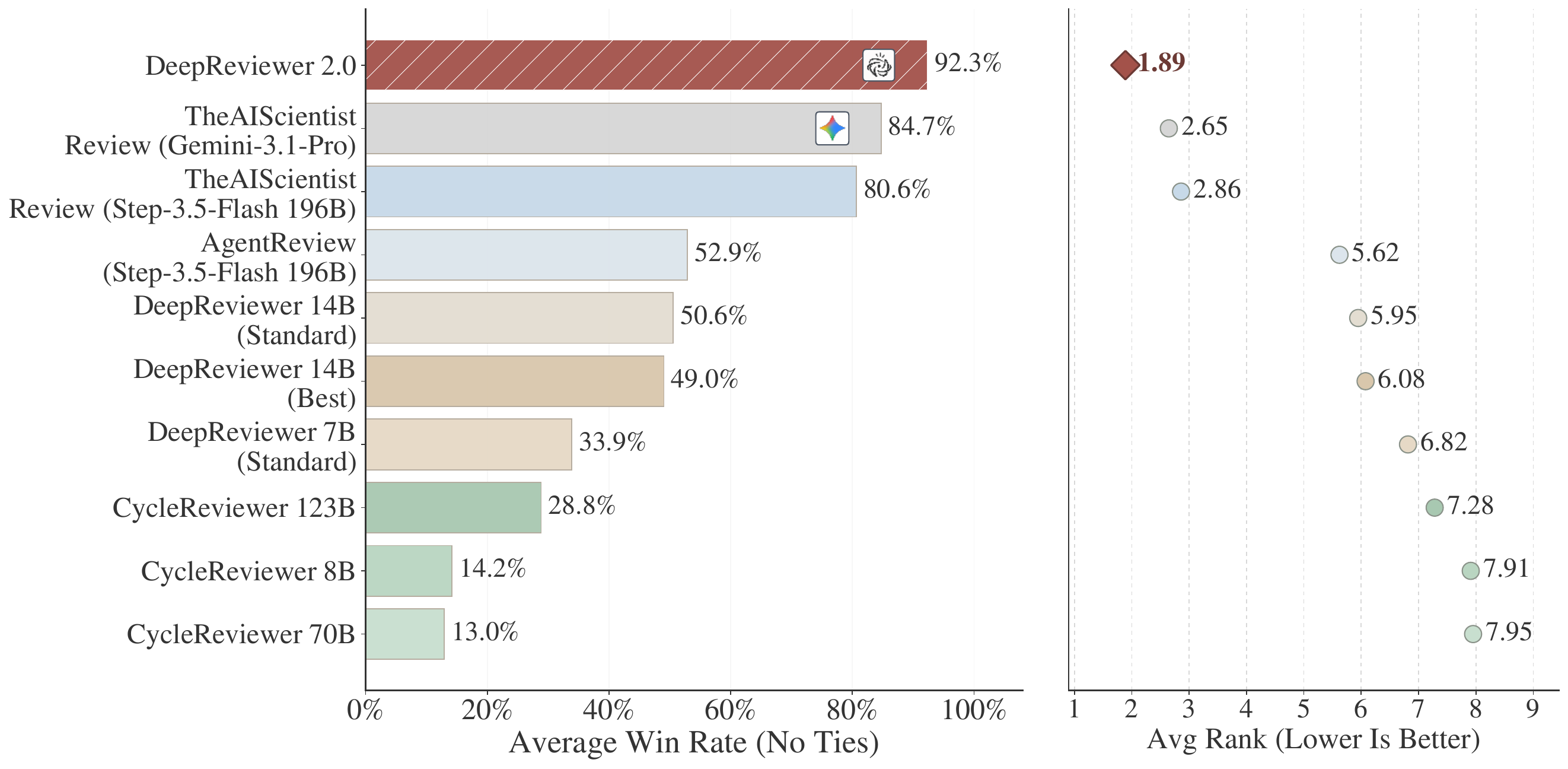}
  \caption{Two-panel summary of anonymous ranking among automatic systems: uniform average non-tie win rate (left) and average rank (right; lower is better).}
  \label{fig:avg-win-rank}
\end{figure}

\textbf{Metric agreement (Figure~\ref{fig:avg-win-rank}).} DeepReviewer~2.0 leads under both uniform average non-tie win rate (92.34\%) and tie-aware average rank (1.889; lower is better), reducing the chance that preference is a single-metric artifact. Taken together, Figures~\ref{fig:pairwise-heatmap}--\ref{fig:avg-win-rank} support a stronger conclusion than ``wins on average'': under a fixed judge and tie rule, DeepReviewer~2.0 is preferred broadly across opponents and remains first under multiple aggregations. We still treat this as judge-conditional evidence of end-to-end usefulness, not as a direct measure of factual correctness.

\paragraph{\textbf{RQ3: In anonymous blind comparison, how does DeepReviewer 2.0 compare to a human review committee?}}
We compare DeepReviewer~2.0 against a human review committee using the same anonymous ranking protocol, and report win/tie/lose percentages in Table~\ref{tab:dr2-vs-human}. On the micro-average across all dimensions, DeepReviewer~2.0 is preferred in 71.63\% of comparisons (ties 16.28\%, losses 12.09\%). The dimension breakdown is more informative than the micro-average alone. The strongest advantages are in Constructive Value (84.50\% win) and Communication Clarity (86.05\% win), consistent with the design goal of producing localized, actionable feedback with an executable repair path. By contrast, margins are smaller in Technical Accuracy (59.69\%) and Analytical Depth (58.14\%), suggesting that process constraints improve evidence organization and revision guidance more than they solve all technical reasoning errors. We therefore interpret RQ3 as a preference signal under this rubric and judge setting, not as a claim of general human replacement.
\input{tables/table_dr2_vs_human}

\paragraph{\textbf{RQ4: What does the category-level coverage profile reveal about strengths and gaps?}}
Figure~\ref{fig:category-coverage} breaks strict coverage down into eight issue categories and shows that DeepReviewer~2.0 is not uniformly best in every category. It leads in 4/8 categories (Novelty, Clarity, Presentation, and Other), while other systems remain stronger in Soundness/Experiments and in ethics-sensitive checks. This profile also explains why a system can lead overall without being best everywhere: the canonical issue list is highly imbalanced (e.g., Experiments has 225 issues and Soundness has 138 issues, while Ethics has only 6 issues). DeepReviewer~2.0's overall advantage therefore comes from concentrated strengths in high-frequency categories together with competitive (though not always best) coverage elsewhere.

The most salient blind spot is Ethics: DeepReviewer~2.0 scores 0.00\% strict coverage there under this protocol, while another system reaches 50.00\%. Even though the category is small in the current benchmark, it matters disproportionately for deployment risk. We therefore read Figure~\ref{fig:category-coverage} as both an explanation of where the current gains come from and a concrete target for improvement: without explicit objectives and dedicated evaluation pressure, ethics-sensitive critique is easy to underweight.

\begin{center}
  \begin{minipage}{\linewidth}
    \centering
    \includegraphics[width=\textwidth]{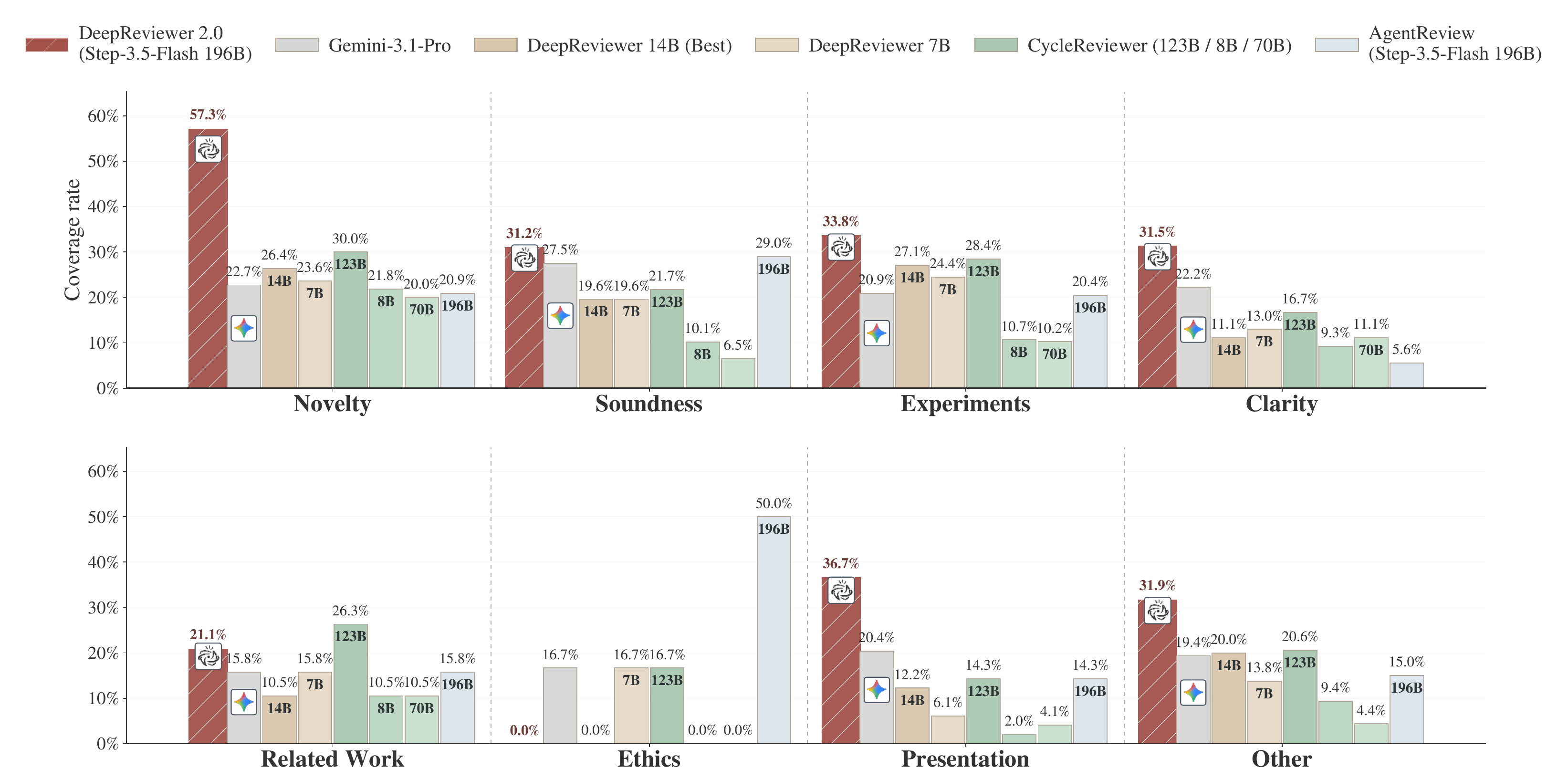}
    \captionof{figure}{Category-level strict issue coverage across eight issue categories.}
    \label{fig:category-coverage}
  \end{minipage}
\end{center}

%% file: tables/table_issue_coverage.tex
\begin{table*}[t]
\centering

\definecolor{drTwoWinBg}{HTML}{F4E8C6}
\resizebox{\textwidth}{!}{%
\begin{tabular}{lcccc}
\toprule
System & Overall (\%) & Major (\%) & Minor (\%) & Critical miss (\%) \\
\midrule
\rowcolor{drTwoWinBg}
DeepReviewer 2.0 (Step-3.5-Flash 196B) & \textbf{35.74} & \textbf{37.26} & \textbf{30.98} & \textbf{62.74} \\

CycleReviewer 123B & 23.92 & 23.74 & 24.46 & 76.26 \\
TheAIScientist Review (Gemini-3.1-Pro) & 21.94 & 23.57 & 16.85 & 76.43 \\
DeepReviewer 14B (Best) & 21.42 & 23.40 & 15.22 & 76.60 \\
AgentReview (Step-3.5-Flash 196B) & 19.58 & 20.80 & 15.76 & 79.20 \\
DeepReviewer 7B (Standard) & 18.92 & 19.58 & 16.85 & 80.42 \\
CycleReviewer 8B & 11.17 & 12.48 & 7.07 & 87.52 \\
CycleReviewer 70B & 9.33 & 10.75 & 4.89 & 89.25 \\
\bottomrule
\end{tabular}
}
\caption{Strict issue coverage against human-identified issues on the evaluation set. Missing system outputs are counted as uncovered. Higher is better for Overall/Major/Minor; lower is better for Critical miss rate ($1-\text{Major}$).}
\label{tab:strict-coverage}
\end{table*}

%% file: tables/table_dr2_vs_human.tex
\begin{table}[!htbp]
\centering
\caption{Anonymous blind preference comparison between DeepReviewer 2.0 and a human review committee (win/tie/lose percentages).}
\label{tab:dr2-vs-human}
\definecolor{drTwoWinBg}{HTML}{F4E8C6}
\begin{tabular}{l >{\columncolor{drTwoWinBg}\bfseries}c c c}
\toprule
Dimension & DeepReviewer 2.0 win (\%) & Tie (\%) & Human win (\%) \\
\midrule
All Dimensions (micro) & 71.63 & 16.28 & 12.09 \\
Technical Accuracy & 59.69 & 24.81 & 15.50 \\
Constructive Value & 84.50 & 5.43 & 10.08 \\
Analytical Depth & 58.14 & 24.03 & 17.83 \\
Communication Clarity & 86.05 & 9.30 & 4.65 \\
Overall Judgment & 69.77 & 17.83 & 12.40 \\
\bottomrule
\end{tabular}
\end{table}

%% file: sections/conclusion.tex
DeepReviewer 2.0 argues that automated reviewing should be judged less by how persuasive it sounds and more by whether its judgments are checkable. We operationalize this stance with a \textbf{review-package} interface and a staged cognitive chain that makes traceability budgets and gated synthesis part of the output contract. Across strict issue coverage, anonymous system ranking, and a blind comparison against a human review committee, the resulting packages are preferred under our protocol. At the same time, strict coverage remains far from complete and ethics coverage can fail sharply; DeepReviewer 2.0 is therefore best read as an \textbf{assistive reviewer} whose value depends on how it is integrated with human oversight and venue-specific checklists.

%% file: sections/limitations.tex
DeepReviewer~2.0 is constrained by design, and some constraints are the point of the paper. The system provides \textbf{process-level} guarantees (structured outputs and traceability links), not semantic truth guarantees: a review package can be coherent and still be wrong about novelty, factual claims, or scientific priorities. This matters because strict issue coverage remains incomplete---even under our protocol, major issues can be missed---so DeepReviewer~2.0 should be used as an assistive tool to augment human reviewers rather than to replace them \citep{Akella2025PrereviewTP}.

Our evaluation protocols also define what the results do and do not measure. Strict issue coverage is computed against a canonical issue list extracted from human reviews/discussion plus the meta-review/decision, which rewards matching what humans already articulated but does not credit newly discovered issues. The anonymous ranking protocols inherit the judge model, rubric, and tie rule; we therefore report multiple complementary aggregates (pairwise win fractions, tie-aware average rank, and an order-invariant Bradley--Terry strength estimate with bootstrap intervals), but different judges could shift preferences, especially on technical accuracy \citep{Kirtani2025ReviewEvalAE,Chu2024PREAP}. Because our judge is Gemini-family while the comparison pool includes a Gemini-3.1-Pro baseline, preferences may also reflect stylistic affinity in addition to substantive quality.

In addition, our quantitative results are conditional on a fixed evaluation pool (Appendix~\ref{app:notation}) that excludes malformed exports and unparseable ranking outputs. This simplifies protocol execution and metric computation, but it can understate end-to-end reliability failures (e.g., tool failures, parser brittleness, or non-executable packages) that matter in unconstrained deployment.

\textbf{Ethical and deployment risks.} Automated reviews can be misused as gatekeeping signals or decision proxies, especially when outputs sound confident but remain fallible. The system is also exposed to manipulation, including hidden-content prompt injection in PDFs \citep{Ye2024AreWT}, and to privacy and data-handling concerns when manuscripts are routed through toolchains for search, retrieval, and rendering. Our category analysis further reveals a sharp gap on Ethics (Figure~\ref{fig:category-coverage}), suggesting that ethics-sensitive critique is easy to underweight when technical content dominates unless it is explicitly pressured by objectives and evaluation. A conservative stance is to treat high-stakes or normatively loaded domains (e.g., medical, legal, dual-use AI) as out of scope unless dedicated safeguards are added on top of technical critique, such as PDF sanitization, explicit ethics checklists, and human oversight with domain expertise.

%% file: sections/appendix.tex
\section{Evaluation instruction templates}
\label{app:prompts}

This appendix summarizes the instruction templates used by the automated judges in Step-2 (strict issue coverage) and Step-3 (anonymous ranking). We state them as evaluation contracts (inputs, decision rules, outputs). The intent is not to imply that results depend on phrasing tricks, but to make the protocol concrete enough to reproduce.

\subsection{Strict issue-coverage evaluator (Step-2)}

\paragraph{Inputs.}
The evaluator receives (i) a canonical issue list extracted from human reviews and (ii) one system-generated review text for the same paper.

\paragraph{Decision rule.}
Each issue is judged independently with a strict binary criterion. The evaluator assigns label $=1$ only when the system review clearly covers the same issue with concrete semantic overlap; label $=0$ is used when the issue is missing, too vague, or only weakly related. External knowledge is disallowed: decisions must rely only on the issue description and the system review text.

\paragraph{Output contract.}
The evaluator returns a single structured record with one entry per issue (issue id, binary label, short reason, optional evidence span from the system review), plus a brief overall assessment. Hard constraints include: every issue id appears exactly once; labels are in $\{0,1\}$; partially related mentions do not count as coverage unless the core concern is addressed.

\subsection{Anonymous ranking judge (Step-3)}

\paragraph{Inputs and anonymity.}
The judge compares multiple candidate reviews for the same paper, treating every candidate as an automated-system output. Candidate identities are anonymized, and the judge is instructed not to use prior assumptions about model family, style, or authorship.

\paragraph{Weakness-driven comparison.}
The judge reads the provided paper text and candidate reviews, then conducts a weakness-by-weakness analysis: it first builds a canonical weakness inventory from the paper and all candidates (splitting fundamental root-cause issues from secondary issues), then maps each candidate's weakness claims to that inventory and assesses correctness, materiality, evidence quality, repairability, and actionability. The judge focuses on substantive review quality and is instructed to ignore minor formatting or surface-level artifacts unless they change meaning.

\paragraph{Ranking chains and tie rule.}
The judge outputs five aspect-specific ranking chains (Technical Accuracy, Constructive Value, Analytical Depth, Communication Clarity, Overall Judgment) using only ``$>$'' and ``$=$''. Ties are governed by an explicit conversion rule: the judge uses internal integer scores in $[1,10]$; score gaps $\le 1$ must be rendered as ties ``$=$'', and only gaps $\ge 2$ may produce strict ``$>$'' relations.

\section{Evaluation metrics and computation details}
\label{app:metrics}

This appendix spells out the metric definitions and computation rules used in the strict-coverage (Step-2) and anonymous-ranking (Step-3) protocols.

\subsection{Notation and dataset filtering}
\label{app:notation}

We evaluate on a sample of 134 ICLR~2025 submissions. Let $\mathcal{P}$ denote the set of evaluated papers, so $|\mathcal{P}|=134$. The ranking-based protocols use a fixed evaluation pool $\mathcal{P}_{\mathrm{rank}}$ consisting of the subset of $\mathcal{P}$ whose Step-3 judge outputs are parseable, and we use the same $\mathcal{P}_{\mathrm{rank}}$ pool for anonymous ranking among systems and for the blind comparison against a human review committee.

For each paper $p \in \mathcal{P}$, Step-1 yields a canonical issue list $\mathcal{I}_p$, where each issue has (i) a severity tag (major/minor) and (ii) a single category from an eight-way taxonomy.

\subsection{Step-2 coverage metrics (strict policy)}
\label{app:coverage-metrics}

For a system $s$, paper $p$, and issue $i \in \mathcal{I}_p$, Step-2 produces a label
\begin{equation}
  y_{s,p,i} \in \{0, 1, \varnothing\},
\end{equation}
where $\varnothing$ denotes missing or unparseable output. Under the strict policy, $\varnothing$ is treated as uncovered.

\paragraph{Overall coverage.}
\begin{equation}
  \mathrm{OverallCoverage}_s
  =
  \frac{\sum_{p \in \mathcal{P}}\ \sum_{i \in \mathcal{I}_p}\mathbf{1}[y_{s,p,i}=1]}
  {\sum_{p \in \mathcal{P}} |\mathcal{I}_p|}.
\end{equation}
\noindent Missing/unparseable output ($\varnothing$) is counted as uncovered under the strict policy.

\paragraph{Severity-split coverage and critical miss rate.}
Let $\mathcal{I}^{\mathrm{major}}_p$ and $\mathcal{I}^{\mathrm{minor}}_p$ be the major and minor issue subsets for paper $p$.
\begin{align}
  \mathrm{MajorCoverage}_s
  &=
  \frac{\sum_{p}\ \sum_{i \in \mathcal{I}^{\mathrm{major}}_p}\mathbf{1}[y_{s,p,i}=1]}
  {\sum_{p} |\mathcal{I}^{\mathrm{major}}_p|},\\
  \mathrm{MinorCoverage}_s
  &=
  \frac{\sum_{p}\ \sum_{i \in \mathcal{I}^{\mathrm{minor}}_p}\mathbf{1}[y_{s,p,i}=1]}
  {\sum_{p} |\mathcal{I}^{\mathrm{minor}}_p|},\\
  \mathrm{CriticalMissRate}_s
  &= 1 - \mathrm{MajorCoverage}_s.
\end{align}

\paragraph{Category coverage.}
For category $c$, let $\mathcal{I}_{p,c}$ denote the issues in paper $p$ that fall into category $c$. We compute
\begin{equation}
  \mathrm{CatCoverage}_{s,c}
  =
  \frac{\sum_{p}\ \sum_{i \in \mathcal{I}_{p,c}}\mathbf{1}[y_{s,p,i}=1]}
  {\sum_{p} |\mathcal{I}_{p,c}|}.
\end{equation}

\subsection{Step-3 ranking metrics}
\label{app:ranking-metrics}

For each paper $p$ and aspect $a$, the judge returns a total preorder over systems, represented as a chain with operators ``$>$'' and ``$=$''. A tie tier is a maximal contiguous group linked by ``$=$''.

\paragraph{Average rank with tie tiers.}
Suppose system $s$ appears in a tie tier occupying positions $r,\ldots,r+k-1$ in the chain (lower is better). We assign the tier-average rank
\begin{equation}
  \mathrm{Rank}_{p,a}(s) = \frac{r+(r+k-1)}{2},
\end{equation}
and define $\mathrm{AvgRank}_s$ as the mean of $\mathrm{Rank}_{p,a}(s)$ over all valid $(p,a)$ instances.

\paragraph{Pairwise non-tie win fraction.}
Let $W_{x,y}$ be the count of strict wins of system $x$ over $y$ aggregated over papers and aspects, counting only ``$>$'' relations and excluding ties. The pairwise non-tie win fraction is
\begin{equation}
  \mathrm{WinFrac}_{x,y}=\frac{W_{x,y}}{W_{x,y}+W_{y,x}}.
\end{equation}
We exclude ties here because, under the judge protocol, ties are induced by score gaps within the judge's 1--10 scoring resolution (gap $\le 1$), and treating them as half-wins can blur decisive preference signals.

\paragraph{Uniform average win rate.}
Let $\mathcal{B}_x$ be the set of opponents of $x$ in the compared pool. The uniform average win rate is
\begin{equation}
  \mathrm{AvgWinRate}_x
  =
  \frac{1}{|\mathcal{B}_x|}\sum_{y\in\mathcal{B}_x}\mathrm{WinFrac}_{x,y}.
\end{equation}

\subsection{Bradley--Terry Elo estimation and bootstrap confidence intervals}
\label{app:bt}

We do not use an online Elo update. Instead, we treat non-tie comparisons as outcomes of a Bradley--Terry model with latent abilities $\{\theta_s\}_{s}$:
\begin{equation}
  \Pr(x \succ y)=\frac{\theta_x}{\theta_x+\theta_y}.
\end{equation}
Let $W_{x,y}$ be the non-tie win counts as above, and let $N_{x,y}=W_{x,y}+W_{y,x}$. We estimate $\theta$ by maximum likelihood and solve it using a standard minorization--maximization (MM) iteration:
\begin{equation}
  \theta_x^{(t+1)}=
  \frac{\sum_{y \ne x} W_{x,y}}
  {\sum_{y \ne x} \frac{N_{x,y}}{\theta_x^{(t)}+\theta_y^{(t)}}}.
\end{equation}
To report results on the familiar Elo scale, we map
\begin{equation}
  \mathrm{Elo}_s = 1500 + 400\log_{10}(\theta_s),
\end{equation}
where we normalize $\theta$ by a constant factor so that the geometric mean ability is 1 (this fixes the BT scale ambiguity without changing relative strengths).

\paragraph{Bootstrap confidence intervals.}
We compute 95\% confidence intervals by paper-level bootstrap with 1000 resamples (sampling papers with replacement). For each resample, we rebuild the win matrix, re-fit the BT model, re-compute Elo, and report the percentile interval $[2.5\%,97.5\%]$.

\subsection{DeepReviewer 2.0 vs human committee percentages}
\label{app:human-percent}

For each aspect $a$, we compare DeepReviewer 2.0 against the human review committee by their ranks in the judge chain on each paper. Let $w_a$, $t_a$, and $\ell_a$ be the counts over papers where $\mathrm{Rank}(\text{DeepReviewer 2.0}) < \mathrm{Rank}(\text{Human})$, equal, or greater, respectively. The reported percentages are normalized per row:
\begin{equation}
  \mathrm{WinPct}_a=\frac{w_a}{w_a+t_a+\ell_a},\quad
  \mathrm{TiePct}_a=\frac{t_a}{w_a+t_a+\ell_a},\quad
  \mathrm{LosePct}_a=\frac{\ell_a}{w_a+t_a+\ell_a}.
\end{equation}

\paragraph{Micro-average across dimensions.}
The ``All Dimensions (micro)'' row aggregates counts across aspects:
\begin{align}
  \mathrm{WinPct}_{\mathrm{micro}}
  &=
  \frac{\sum_a w_a}{\sum_a (w_a+t_a+\ell_a)},\\
  \mathrm{TiePct}_{\mathrm{micro}}
  &=
  \frac{\sum_a t_a}{\sum_a (w_a+t_a+\ell_a)},\\
  \mathrm{LosePct}_{\mathrm{micro}}
  &=
  \frac{\sum_a \ell_a}{\sum_a (w_a+t_a+\ell_a)}.
\end{align}

\section{Document parsing and anchored annotation}
\label{app:anchored-annotation}

This appendix clarifies how DeepReviewer 2.0 turns a PDF into a traceable text representation and how paragraph-level feedback is rendered as anchored annotations. We focus on the representation contract: what information is carried forward so that every critique can be traced back to a concrete page region, even when the input PDF is noisy.

\subsection{From PDF to page-indexed semantic text}
\label{app:pdf-to-text}

DeepReviewer 2.0 normalizes each input PDF into (i) a page-indexed semantic text stream and (ii) a layout trace produced by the PDF parser. In our implementation, we use MinerU (v4) to produce markdown and a \texttt{content\_list} of extracted blocks; each block records its page index, text, type (e.g., header/text/equation/table/image), and (when available) a bounding box. We then build a page index by ordering the extracted texts per page:
\begin{equation}
  \mathcal{J}(p) = [t_{p,1}, t_{p,2}, \ldots, t_{p,L_p}],
\end{equation}
where $L_p$ is the number of extracted lines on page $p$. When layout coordinates are available, each line $t_{p,k}$ additionally carries a bounding box $\mathrm{BBox}(p,k)=[x_{\min},y_{\min},x_{\max},y_{\max}]$ together with per-page reference scales $(W_p,H_p)$ from the parser output. This dual view is deliberate: the text stream supports reasoning and comparison, while the optional boxes support rendering highlights; when boxes are missing or noisy, traceability still holds through stable page-and-line anchors.
Coordinates follow a top-left origin convention with $x$ increasing rightward and $y$ increasing downward.

\begin{figure*}[t]
  \centering
  \includegraphics[width=\textwidth]{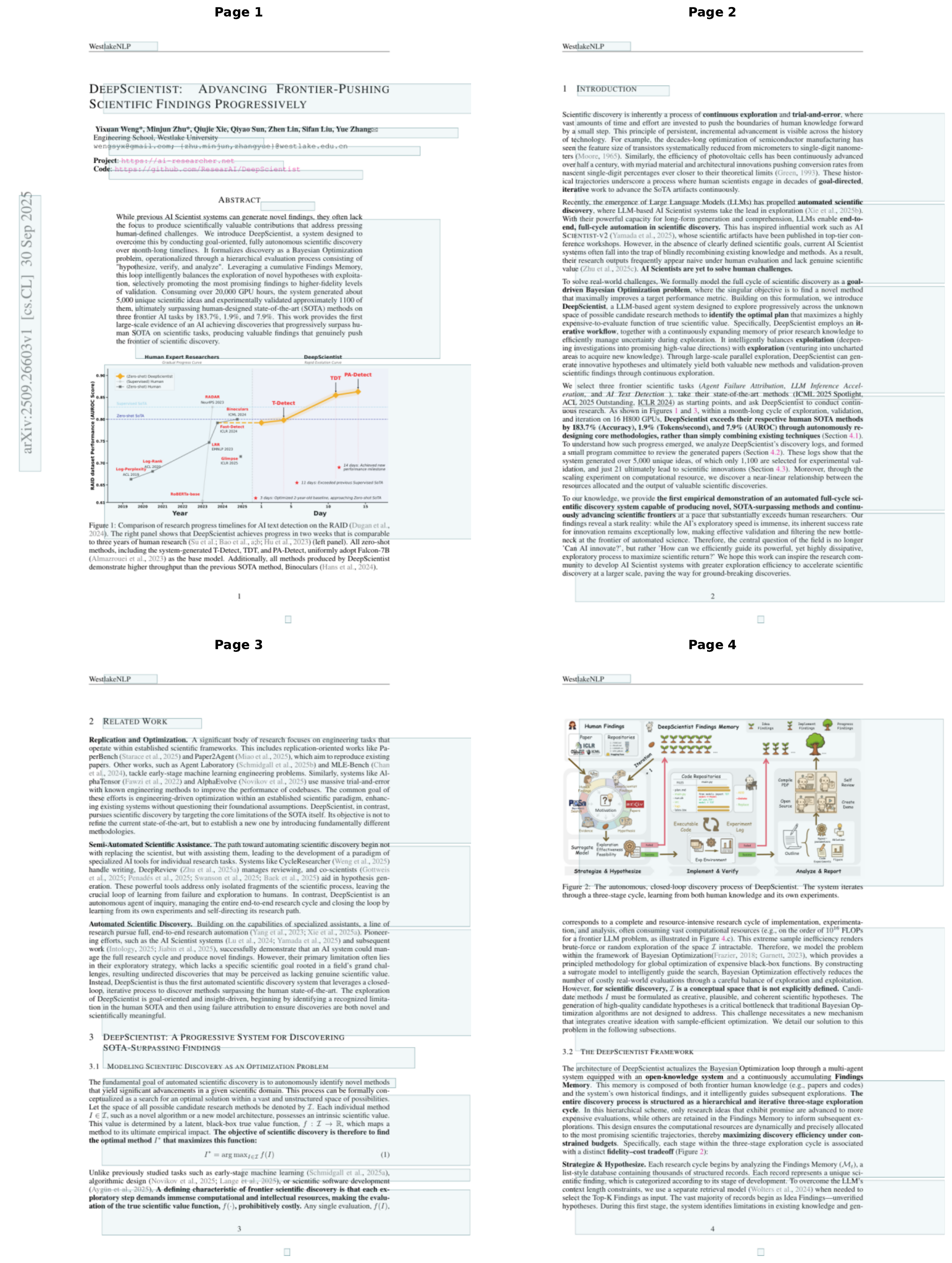}
  \caption{MinerU parsing visualization on the first four pages of \texttt{arXiv:2509.26603}. Semi-transparent rectangles show the block-level bounding boxes returned in \texttt{content\_list}. DeepReviewer~2.0 builds the page-indexed text stream $\mathcal{J}(p)$ by ordering these blocks per page, enabling stable anchors of the form $(p,k^{\text{start}},k^{\text{end}})$ and later overlay rendering.}
  \label{fig:mineru-parsing-demo}
\end{figure*}

In practice, the parser returns bounding boxes in a page-local coordinate system with an implicit reference scale. We treat each box as $b_{p,k}=[x_{\min},y_{\min},x_{\max},y_{\max}]$ together with a per-page reference $(W_p,H_p)$ (computed from the maximum observed extents on that page), so that we can normalize by $(W_p,H_p)$ and map highlights back onto the source PDF consistently during export (Section~\ref{app:annotation-objects}). Figure~\ref{fig:mineru-parsing-demo} visualizes this parsing contract on a real paper: each semi-transparent rectangle is a MinerU-returned block box on the first four pages of \texttt{arXiv:2509.26603}.

\subsection{Implementation pseudocode (MCP tool loop)}
\label{app:pseudocode}

Figure~\ref{fig:dr2-pipeline} summarizes the end-to-end pipeline as executed in our backend. The key point is that ``real-world'' interactions are not implicit: the agent reads pages and writes annotations through MCP tools, and the export gate is checked over explicit counters and artifacts (Section~\ref{sec:method}).

\begin{figure}[t]
  \centering
  \begin{minipage}{0.98\linewidth}
    \footnotesize
\begin{verbatim}
Algorithm 1: DeepReviewer 2.0 pipeline (implementation view)
Input : source-paper PDF D
Output: review package Y=(R, A, P, N)

1  (md, content_list) <- MinerUParse(D)
2  Build page-indexed stream J(p) and optional BBox(p,k) from content_list
3  L^(0) <- f_meta(J)                         # Stage I: independent pre-review
4  Q <- g(L^(0))                              # investigation agenda
5  for q in Q do
6    C_q <- paper_search(q)                   # MCP tool; PASA-backed
7    q_hat <- VerifyComparable(C_q)           # matched-setting gate
8    Update ledger L and novelty tags N using q_hat
9  end for
10 for prioritized anchor span (p,k_start,k_end) do
11   pdf_read_lines(p,k_start,k_end)          # MCP tool; evidence verification
12   A <- A U {pdf_annotate(p,k_start,k_end,tau,sigma,rho,delta)}  # MCP tool
13 end for
14 Enforce export gate (Eq. 8) on schema, budgets, and |A|
15 R <- review_final_markdown_write(...)      # MCP tool; sectioned writes
16 Export final_report.pdf by overlaying A on appended source pages
17 return Y
\end{verbatim}
  \end{minipage}
  \caption{End-to-end DeepReviewer 2.0 pipeline in an implementation-oriented view. Tool names correspond to MCP endpoints used by the agent loop (Section~\ref{sec:method}).}
  \label{fig:dr2-pipeline}
\end{figure}

\subsection{Annotation objects and anchor resolution}
\label{app:annotation-objects}

Each annotation couples a location anchor with a diagnosis and a repair action. In the main paper we write an annotation as $a_j=(\ell_j,\tau_j,\sigma_j,\rho_j,\delta_j)$ (Equation~\ref{eq:annotation-unit}). Operationally, the location $\ell_j$ is a page and line span:
\begin{equation}
  \ell_j = (p_j, k_j^{\text{start}}, k_j^{\text{end}}), \qquad 1 \le k_j^{\text{start}} \le k_j^{\text{end}} \le L_{p_j}.
\end{equation}
The annotation also carries a short summary and a longer comment text used for the callout body. For example, an annotation may target page 2, lines 14--22, be marked as a major issue, summarize an overstated novelty claim, explain the scientific risk, and then specify a concrete revision requirement (often including a proposed rewrite).

To render a highlight overlay, we convert $(p_j,k_j^{\text{start}},k_j^{\text{end}})$ into a set of rectangles on the source PDF page. When parser bounding boxes are available, we take the union of the boxes covered by the line range:
\begin{equation}
  \mathcal{R}_j = \bigcup_{k=k_j^{\text{start}}}^{k_j^{\text{end}}} \mathrm{BBox}(p_j,k).
\end{equation}
When bounding boxes are unavailable, we fall back to a line-ratio approximation that allocates a vertical span proportional to the line indices:
\begin{equation}
  y_1 = 100\cdot \frac{k_j^{\text{start}}-1}{L_{p_j}}, \quad
  y_2 = 100\cdot \frac{k_j^{\text{end}}}{L_{p_j}},
\end{equation}
and use a conservative fixed horizontal range (e.g., $x\in[8,92]$ in a normalized 0--100 coordinate system) so that the highlight remains visible even under extraction noise.

\subsection{Overlay rendering and continuation pages}
\label{app:overlay}

DeepReviewer 2.0 exports anchored feedback as a combination of (i) translucent highlights over the source page regions and (ii) right-margin callouts that carry the critique text. Each callout receives a stable marker (page index plus per-page annotation index), and the renderer routes a short connector line from the highlighted region to the corresponding callout while prioritizing non-overlap among callout boxes.

Long comments are common in reviewer-style writing. When a callout cannot fit in the right margin without overlap, DeepReviewer 2.0 switches to a continuation mode: the source page keeps a compact index card indicating that the full text is moved, and the complete comment is placed on a continuation sheet linked back to the source highlight. This yields an annotated PDF where the source page remains readable while the critique retains enough detail to be executable.

\section{Case study: an anchored review package}
\label{app:case-study}

We include a representative case study to make the review package interface concrete. The example comes from a generated review of an arXiv preprint (\texttt{arXiv:2509.26603}). We reproduce (i) two four-page excerpts from the exported PDF that illustrate anchored highlights and continuation sheets, and (ii) the complete structured review report.

\begin{center}
  \begin{minipage}{\textwidth}
    \centering
    \includegraphics[width=\textwidth]{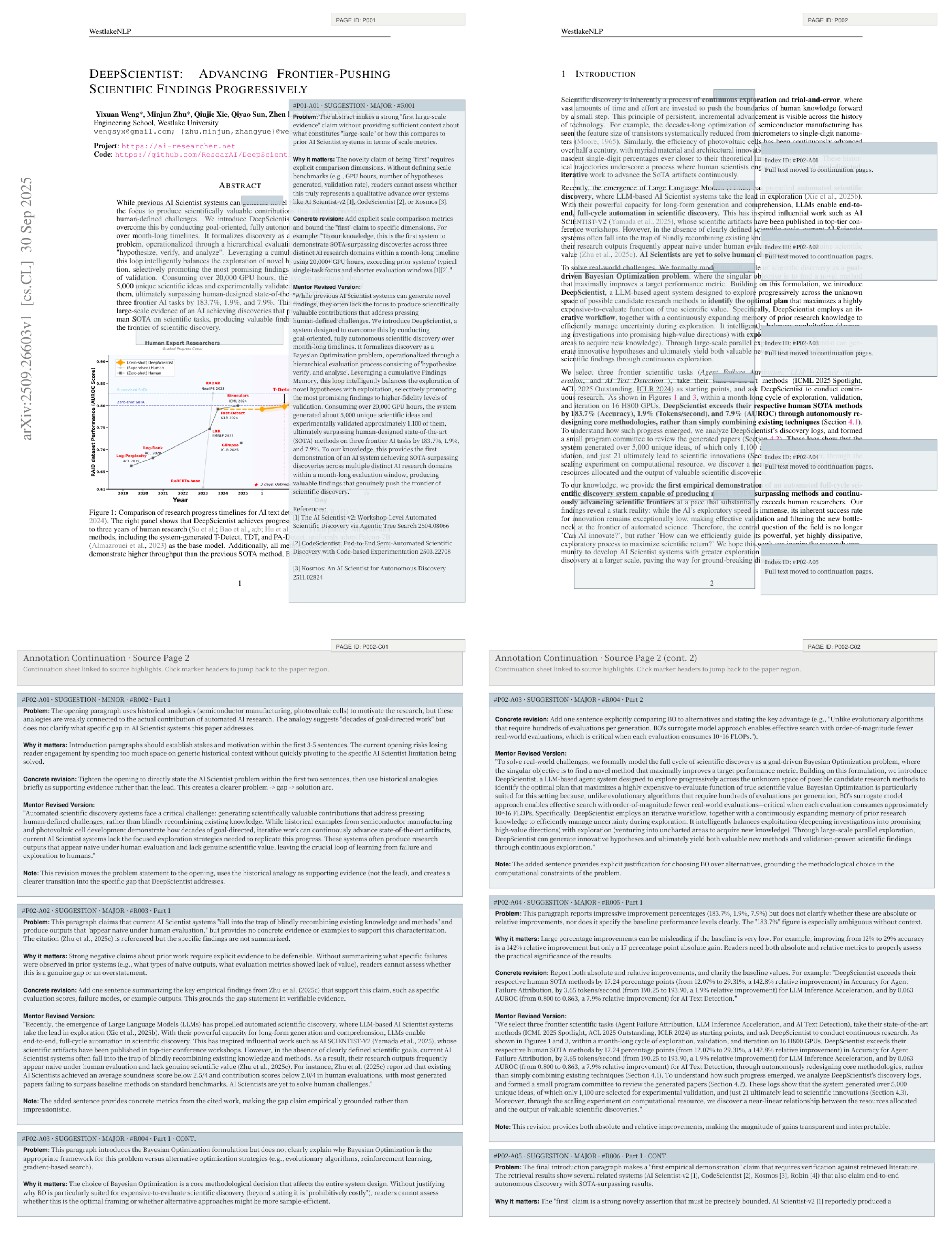}
    \captionof{figure}{Excerpt (4 consecutive pages) from an exported review-package PDF (2$\times$2 collage). Pages preserve the original paper layout and include translucent highlights plus margin callouts; when a callout cannot fit cleanly, the exporter inserts continuation sheets linked to the source highlight.}
    \label{fig:case-report-collage}
  \end{minipage}
\end{center}

\begin{center}
  \begin{minipage}{\textwidth}
    \centering
    \includegraphics[width=\textwidth]{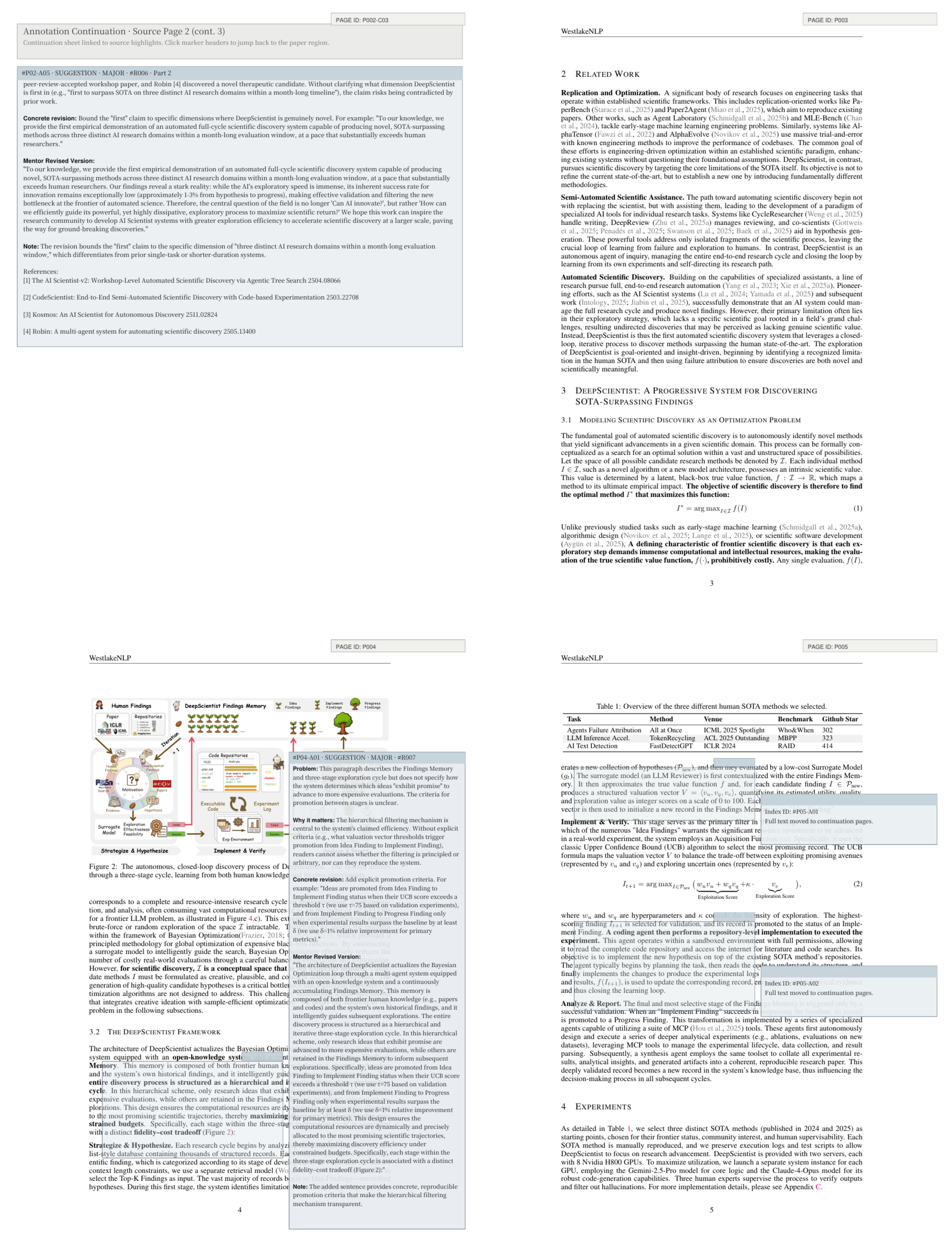}
    \captionof{figure}{Additional excerpt (4 consecutive pages) from the same exported review-package PDF (2$\times$2 collage). This span illustrates how anchored critiques can extend across sections while keeping the source pages readable and the repair actions explicit.}
    \label{fig:case-report-collage-2}
  \end{minipage}
\end{center}

The full report illustrates the global layer: a structured summary followed by strengths, weaknesses, and a decision recommendation. This is where DeepReviewer 2.0 tries to stay honest about what it does and does not know. It is tempting for automated systems to write confident prose; the report format forces a more checkable decomposition (e.g., explicitly separated novelty concerns, methodological transparency concerns, and statistical reporting concerns). In this case, the global judgment depends heavily on whether the paper's ``first'' claims survive direct comparison to closely related autonomous-scientist systems under matched settings. Instead of leaving novelty as an impressionistic statement, the report records concrete comparators and qualifies verdicts when direct comparability is unclear.

Figures~\ref{fig:case-report-collage} and \ref{fig:case-report-collage-2} illustrate the local layer: each callout follows a repeatable structure that mirrors our cognitive-chain design. A paragraph role diagnosis explains what that passage is supposed to do in the paper; a concrete defect states what is wrong; a risk paragraph explains why it matters scientifically; and a revision requirement gives a specific action the authors can execute, often accompanied by a proposed rewrite. The continuation sheets make this format practical: the source page stays readable, but the critique retains enough detail to be actionable. This is the core interface claim of DeepReviewer 2.0: we are not trying to ``sound like a reviewer''; we are trying to leave behind a trail that another human can check, disagree with, and still use to revise the paper.

\clearpage


\section*{Supplementary material}

# Final Agent Report

Structured markdown synthesized by the annotation agent.

## Summary

This paper presents DeepScientist, an autonomous AI system for scientific discovery that formalizes the research process as a Bayesian Optimization problem. The system employs a three-stage hierarchical exploration cycle (Strategize & Hypothesize, Implement & Verify, Analyze & Report) with a cumulative Findings Memory to balance exploration and exploitation. Evaluated on three frontier AI tasks (Agent Failure Attribution, LLM Inference Acceleration, AI Text Detection), DeepScientist reportedly surpasses human SOTA methods by 183.7%, 1.9%, and 7.9% respectively, using 20,000+ GPU hours over a month-long timeline.

**Strengths:** The paper addresses an important and timely problem—improving the scientific value of AI-generated research. The Bayesian Optimization framing is conceptually sound for expensive-to-evaluate search spaces. The hierarchical filtering mechanism with Findings Memory represents a thoughtful approach to sample efficiency. The empirical evaluation across three distinct domains demonstrates system versatility.

**Weaknesses:** Several critical issues undermine confidence in the claims. The "first large-scale evidence" novelty claim lacks precise bounding against prior systems like AI Scientist-v2, CodeScientist, and Kosmos that also demonstrate end-to-end autonomous discovery. Key methodological details are missing: promotion criteria between exploration stages, operational definitions for the LLM Reviewer's valuation components (utility, quality, exploration value), and failure handling policies for the 50% implementation failure rate. The reported percentage improvements (especially 183.7%) are presented without clarifying absolute vs. relative gains, which can be misleading when baselines are low. The human evaluation protocol, while emulating ICLR standards, lacks details on reviewer independence and potential conflicts of interest given the author-affiliated program committee.

**Key Issues:** (1) Novelty claims require precise dimension bounding to differentiate from prior work; (2) Method reproducibility is compromised by missing operational definitions; (3) Result transparency requires both absolute and relative improvement metrics; (4) Causal attribution claims need ablation controls to isolate mechanism effects from confounders.

**Recommendation:** Major revisions required before acceptance. The core contribution is valuable, but claims must be precisely bounded, methodological details completed for reproducibility, and causal claims supported by appropriate controls.

## Strengths

1. **Important Problem Addressed:** The paper tackles a critical bottleneck in AI-driven scientific discovery—moving beyond naive recombination of existing knowledge to produce genuinely valuable scientific contributions. This is a timely and significant challenge as AI Scientist systems become more prevalent.

2. **Conceptually Sound Framework:** The Bayesian Optimization formulation for expensive-to-evaluate scientific discovery is well-motivated. The three-stage hierarchical

exploration cycle (Strategize & Hypothesize, Implement & Verify, Analyze & Report) with Findings Memory represents a thoughtful approach to balancing exploration and exploitation under computational constraints.

3. **Comprehensive Empirical Evaluation:** The system is evaluated across three distinct AI research domains (Agent Failure Attribution, LLM Inference Acceleration, AI Text Detection), demonstrating versatility. The reported improvements over human SOTA methods, particularly the 183.7% relative gain in Agent Failure Attribution accuracy, suggest meaningful progress.

4. **Large-Scale Experimentation:** The month-long evaluation timeline with 20,000+ GPU hours, 5,000 generated ideas, and 1,100 validated hypotheses represents one of the most extensive empirical demonstrations of autonomous scientific discovery to date.

5. **Human Evaluation Protocol:** The paper includes both automated evaluation (DeepReviewer) and human expert evaluation emulating ICLR standards, with reported inter-rater reliability (Krippendorff's $\alpha = 0.739$). The average human rating of 5.00 closely matches the ICLR 2025 average of 5.08.

6. **Scaling Analysis:** The near-linear relationship between computational resources and discovery output (Figure 6) provides valuable insights for future system design and resource allocation.

7. **Ethical Considerations:** The paper includes a thoughtful Ethics Statement addressing dual-use risks and academic ecosystem impacts, with concrete mitigation measures including selective open-sourcing and mandatory human supervision requirements.

## Weaknesses

### 1. Novelty Claims Lack Precise Bounding (Critical)

The paper makes strong "first" claims without adequately differentiating from prior AI Scientist systems:

- **Abstract claim:** "first large-scale evidence of an AI achieving discoveries that progressively surpass human SOTA"
- **Introduction claim:** "first empirical demonstration of an automated full-cycle scientific discovery system capable of producing novel, SOTA-surpassing methods"

However, retrieved literature shows several related systems with comparable claims:

- AI Scientist-v2 [1] produced a peer-review-accepted workshop paper with end-to-end autonomous discovery
- Robin [2] discovered and validated a novel therapeutic candidate for dAMD
- Kosmos [3] demonstrated linear scaling of discoveries with system cycles
- CodeScientist [4] reported 19 discoveries with 6 judged as sound and incrementally novel

Without bounding the "first" claim to specific dimensions (e.g., "first to surpass SOTA on three distinct AI domains within a month-long timeline"), the novelty assertion risks being contradicted.

### 2. Missing Methodological Details Compromise Reproducibility (Major)

Several critical implementation details are absent:

- **Promotion criteria:** What UCB score threshold triggers promotion from Idea Finding to Implement Finding? What minimum improvement delta triggers Progress Finding status?

- **Valuation vector definitions:** How are utility ($v_u$), quality ($v_q$), and exploration value ($v_e$) operationally defined and computed by the LLM Reviewer?
- **Failure handling:** Appendix C mentions 50% implementation failure rate due to timeouts, but the main text does not describe retry policies or how failed implementations are recorded.
- **UCB hyperparameters:** While Appendix C states $w_u=1$, $w_q=1$, $\kappa=1$, it does not explain how these were selected or whether sensitivity analysis was performed.

3. Result Transparency Issues (Major)

The reported improvements lack clarity:

- **183.7% improvement** in Agent Failure Attribution is a relative gain from 12.07% to 29.31% (only 17.24 percentage points absolute). Presenting only the relative figure without absolute context can mislead readers about practical significance.
- **1.9% improvement** in LLM Inference Acceleration (190.25 to 193.90 tokens/second) is a modest 3.65 tokens/second gain that may not be practically meaningful given the 20,000 GPU hour investment.
- No variance or confidence intervals are reported for any results, making it impossible to assess statistical reliability.

4. Causal Attribution Without Controls (Major)

The paper attributes gains to specific methodological innovations (e.g., A2P's counterfactual reasoning, ACRA's long-term memory) but provides no ablation studies or matched-capacity controls to isolate these effects from confounders like:

- Increased computation per hypothesis
- Different implementation quality
- Hyperparameter tuning differences
- Selection bias in which hypotheses reach validation

5. Human Evaluation Protocol Limitations (Major)

While the paper emulates ICLR review standards:

- The program committee consists of volunteers, but their independence from the authors is not fully established
- No information on whether reviewers knew the papers were AI-generated during initial evaluation
- The 55-minute average review time is notably shorter than typical ICLR reviews, potentially affecting evaluation depth
- Only 5 papers were evaluated, limiting statistical power

6. Overstated Efficiency Claims (Minor)

The paper claims the targeted exploration achieved breakthroughs using 20,000 GPU hours versus 100,000+ for naive testing of all 5,000 candidates. However:

- This comparison assumes all 5,000 ideas would require full evaluation, but the hierarchical filtering means most ideas never reach expensive validation stages
- The actual efficiency gain depends on the counterfactual of how many ideas a random sampling approach would need to test, which is not empirically established

Key Issues

Issue 1: Novelty Claim Overreach

**Location:** Page 1 - Abstract; Page 2 - Introduction (final paragraph)

**Severity:** Critical

**Evidence:** The abstract claims "first large-scale evidence" and introduction claims "first empirical demonstration" without bounding to specific dimensions. Retrieved papers [1-4] demonstrate comparable end-to-end autonomous discovery capabilities.

**Impact:** If uncorrected, reviewers may reject the paper based on perceived novelty overstatement, regardless of technical merit.

**Fix Required:** Bound "first" claims to specific dimensions where DeepScientist is genuinely novel (e.g., "first to demonstrate SOTA-surpassing discoveries across three distinct AI research domains within a month-long evaluation window").

Issue 2: Missing Reproducibility Details

**Location:** Page 4-5 - Method section (Strategize & Hypothesize, Implement & Verify paragraphs)

**Severity:** Major

**Evidence:** No promotion criteria thresholds, no operational definitions for valuation vector components, no failure handling policy described in main text.

**Impact:** Other researchers cannot reproduce the system or fairly compare against it.

**Fix Required:** Add explicit thresholds (e.g., UCB score $\tau$=75 for promotion), define $v_u/v_q/v_e$ computation, describe retry policies and error logging.

Issue 3: Misleading Improvement Metrics

**Location:** Page 2 - Introduction; Page 6 - Experiments

**Severity:** Major

**Evidence:** 183.7% improvement is relative gain from low baseline (12.07% to 29.31%); absolute gain is only 17.24 percentage points. No variance reported.

**Impact:** Readers may overestimate practical significance of results.

**Fix Required:** Report both absolute and relative improvements; add variance/confidence intervals across multiple runs.

Issue 4: Unsupported Causal Attribution

**Location:** Page 6 - Agent Failure Attribution paragraph

**Severity:** Major

**Evidence:** Claims A2P's counterfactual reasoning causes improvements but provides no ablation removing counterfactual component while holding other factors constant.

**Impact:** Cannot determine whether gains come from claimed mechanism or confounders (e.g., implementation quality, hyperparameters).

**Fix Required:** Add matched-capacity ablation studies; use cautious wording ("consistent with") until causal controls are added.

Issue 5: Human Evaluation Independence Unclear

**Location:** Page 7 - Human Expert Evaluation; Appendix A

**Severity:** Major

**Evidence:** Program committee described as "volunteers" but independence from authors not established; 55-minute review time is short; only 5 papers evaluated.

**Impact:** Evaluation credibility depends on reviewer independence and evaluation depth.

**Fix Required:** Clarify reviewer independence, provide reviewer instructions, discuss limitations of small sample size.

## Actionable Suggestions

S1: Bound Novelty Claims with Explicit Dimensions

**Priority:** P0 (Must)

**Action:** Revise abstract and introduction "first" claims to specify exact dimensions of novelty.

**Concrete Revision:** Replace "first large-scale evidence" with "first demonstration of SOTA-surpassing discoveries across three distinct AI research domains within a month-long evaluation window using 20,000+ GPU hours."

**Expected Benefit:** Prevents novelty rejection based on comparison with prior single-task or shorter-duration systems.

S2: Add Method Reproducibility Details

**Priority:** P0 (Must)

**Action:** Add explicit promotion criteria, valuation definitions, and failure handling to Section 3.2.

**Concrete Revision:**

- Add: "Ideas are promoted from Idea Finding to Implement Finding when UCB score exceeds $\tau$=75 (selected via validation experiments), and to Progress Finding only when experimental results surpass baseline by $\delta \geq 1\%$ relative improvement."
- Add: "Utility ($v_u$) measures expected performance improvement based on semantic similarity to prior successful findings; Quality ($v_g$) assesses methodological soundness via structured rubric; Exploration value ($v_e$) quantifies novelty as cosine distance from Findings Memory embeddings."
- Add: "Failed implementations trigger up to 3 automatic retries with debugging; after exhaustion, findings are marked 'Implementation Failed' with error logs retained to inform future hypothesis generation."

  **Expected Benefit:** Enables reproduction and fair comparison by other researchers.

S3: Report Both Absolute and Relative Improvements

**Priority:** P0 (Must)

**Action:** Revise all result statements to include absolute gains alongside relative percentages.

**Concrete Revision:** Replace "183.7% improvement" with "17.24 percentage point improvement (from 12.07% to 29.31%, a 142.8% relative gain)" throughout.

**Expected Benefit:** Provides transparent assessment of practical significance.

S4: Add Variance and Statistical Tests

**Priority:** P1 (High)

**Action:** Run 3-5 additional seeds for each task and report mean±std with significance tests against baselines.

**Concrete Revision:** Add table row showing "DeepScientist (mean±std)" with paired t-test p-values against human SOTA.

**Expected Benefit:** Establishes statistical reliability of reported gains.

S5: Add Matched-Capacity Ablations

**Priority:** P1 (High)

**Action:** For A2P and ACRA, create ablated variants removing claimed core mechanisms while holding parameter count and compute constant.

**Concrete Revision:** Add experiment: "A2P w/o Counterfactual" (uses pattern matching instead of abduction-action-prediction) and "ACRA w/o Memory" (standard decoder without suffix pattern grafting).

**Expected Benefit:** Isolates causal effect of claimed mechanisms from confounders.

S6: Clarify Human Evaluation Independence

**Priority:** P1 (High)

**Action:** Add details to Appendix A about reviewer selection, independence, and instructions.

**Concrete Revision:** Specify: "All three reviewers had no prior collaboration with authors in the past 3 years"; "Reviewers were informed papers were AI-generated after initial scoring to assess bias"; "Full reviewer instructions provided in supplementary material."

**Expected Benefit:** Strengthens credibility of human evaluation results.

S7: Add Discussion of Computational Cost vs. Gain Tradeoff

**Priority:** P2 (Nice-to-have)

**Action:** Add paragraph discussing whether 20,000 GPU hours is justified by observed gains, particularly for the 1.9% LLM acceleration result.

**Concrete Revision:** "While the 1.9% throughput improvement may appear modest relative to computational investment, the scientific contribution lies in demonstrating autonomous methodology redesign rather than engineering optimization. For practical deployment, combining ACRA with established techniques like layer skipping could yield larger gains at lower cost."

**Expected Benefit:** Preempts reviewer concerns about cost-effectiveness.

## Storyline Options + Writing Outlines

### Current Storyline Analysis

The current introduction follows this structure:

· P1: Historical analogies (semiconductor, photovoltaic) motivating iterative advancement
· P2: LLMs enabling AI Scientist systems, but current systems produce naive outputs
· P3: Proposed solution - Bayesian Optimization formulation with DeepScientist
· P4: Three tasks, results preview, scaling findings
· P5: "First" claim, bottleneck identification, call to action

**Alignment Check:**

· (a) Problem-Solution Alignment: Partial - the gap (naive AI Scientist outputs) is stated but not empirically grounded with specific metrics
· (b) Variable Alignment: Weak - core concepts (Findings Memory, UCB, valuation vector) are introduced in method but not foreshadowed in introduction
· (c) Contribution-Evidence Alignment: Partial - claims of SOTA-surpassing are supported by tables, but causal attribution lacks ablation evidence

Alternative Storyline Option A (Problem-First)

**P1 (Problem):** Start with the core challenge: "AI Scientist systems can generate novel findings but struggle to produce scientifically valuable contributions addressing human-defined challenges."

**P2 (Evidence of Gap):** Cite specific evaluation metrics from prior work: "Recent evaluations show AI-generated papers achieving average soundness scores below 2.5/4 and contribution scores below 2.0/4, with most failing to surpass baseline methods."

**P3 (Solution):** Introduce DeepScientist's Bayesian Optimization framing and hierarchical filtering as the response to this gap.

**P4 (Results):** Preview key findings with both absolute and relative improvements.

**P5 (Implications):** Bound novelty claim, identify validation bottleneck, call for exploration efficiency research.

Alternative Storyline Option B (Contribution-First)

**P1 (Contribution):** "We present DeepScientist, the first AI system to demonstrate SOTA-surpassing discoveries across three distinct AI research domains within a month-long evaluation window."

**P2 (Method Insight):** Explain the key innovation: hierarchical exploration with Findings Memory balancing exploration/exploitation.

**P3 (Motivation):** Why this matters: current AI Scientists lack goal-directed focus, producing naive outputs.

**P4 (Evidence):** Results preview with metrics.

**P5 (Broader Impact):** Validation bottleneck, efficiency challenge, call to action.

Recommended Storyline: Option A (Problem-First)

**Rationale:** Option A creates stronger narrative tension by establishing the problem before the solution, which improves reader engagement and makes the contribution feel more necessary.

Abstract Outline (4-5 Sentence Structure)

**S1 (Problem):** "While previous AI Scientist systems can generate novel findings, they often lack the focus to produce scientifically valuable contributions that address pressing human-defined challenges."

**S2 (Gap):** "Current systems frequently produce research outputs that appear naive under human evaluation, with reported soundness scores below 2.5/4 and contribution scores below 2.0/4."

**S3 (Method):** "We introduce DeepScientist, a goal-oriented autonomous discovery system formalizing research as Bayesian Optimization with hierarchical 'hypothesize, verify, analyze' evaluation and cumulative Findings Memory."

**S4 (Results):** "Across 20,000 GPU hours and 5,000 generated ideas, DeepScientist surpassed human SOTA on three AI tasks by 17.24 percentage points (Agent Failure Attribution), 3.65 tokens/second (LLM Inference Acceleration), and 0.063 AUROC (AI Text Detection)."

**S5 (Claim):** "To our knowledge, this demonstrates the first AI system achieving SOTA-surpassing discoveries across three distinct AI research domains within a month-long timeline, revealing validation efficiency as the new frontier bottleneck."

Introduction Outline (Paragraph-by-Paragraph)

**P1 (Problem Statement):**

· Role: Establish the core challenge
· Target Claim: AI Scientists lack focus on scientifically valuable contributions
· Transition: From general AI Scientist promise to specific limitation
· Evidence: Cite Zhu et al. (2025c) evaluation metrics

**P2 (Gap Evidence):**

· Role: Ground the problem in empirical findings
· Target Claim: Prior systems produce naive outputs with low evaluation scores
· Transition: From problem to need for goal-directed approach
· Evidence: Specific scores from retrieved literature

**P3 (Solution - Bayesian Optimization):**

· Role: Introduce the methodological framing
· Target Claim: BO is suited for expensive-to-evaluate scientific discovery
· Transition: From problem to proposed approach
· Evidence: Compare BO sample efficiency to evolutionary alternatives

**P4 (DeepScientist Overview):**

· Role: Present the system
· Target Claim: Hierarchical exploration with Findings Memory enables efficient search
· Transition: From framing to implementation
· Evidence: Preview three-stage cycle

**P5 (Results Preview):**

· Role: Preview empirical findings
· Target Claim: DeepScientist surpasses human SOTA on three tasks
· Transition: From method to results
· Evidence: Both absolute and relative improvements with task names

**P6 (Analysis Findings):**

· Role: Present discovery process insights
· Target Claim: 5,000 ideas → 1,100 validated → 21 innovations; near-linear scaling
· Transition: From results to deeper analysis
· Evidence: Success rates, scaling relationship

**P7 (Bounded Novelty + Bottleneck):**

· Role: State contribution with precise bounds
· Target Claim: First to demonstrate X across Y domains in Z timeline; validation is new bottleneck
· Transition: From findings to implications
· Evidence: Comparison to prior single-task systems

**P8 (Call to Action):**

· Role: Frame future research direction
· Target Claim: Field must shift from "Can AI innovate?" to "How to guide exploration efficiently?"

- Transition: From contribution to community impact
- Evidence: None needed (forward-looking statement)

## Priority Revision Plan

### P0 (Must Fix Before Submission) - Estimated 3-5 days

| Issue | Risk Level | Root Cause | Recommended Fix | Expected Benefit | Priority |
|---|---|---|---|---|---|
| Novelty claim overreach | Critical | Unbounded "first" assertions | Bound claims to specific dimensions (3 domains, month-long timeline) | Prevents novelty-based rejection | P0 |
| Missing promotion criteria | Major | Incomplete method description | Add UCB threshold $\tau=75$, improvement delta $\delta=1\%$ | Enables reproduction | P0 |
| Missing valuation definitions | Major | Opaque LLM Reviewer scoring | Define $v_u$, $v_q$, $v_e$ computation with prompt reference | Enables reproduction | P0 |
| Misleading improvement metrics | Major | Relative-only reporting | Add absolute gains (e.g., 17.24 pp, not just 183.7%) | Transparent significance | P0 |
| Missing failure handling | Major | Incomplete implementation description | Add retry policy (3 attempts), error logging | Completes method description | P0 |

### P1 (High Priority) - Estimated 1-2 weeks

| Issue | Risk Level | Root Cause | Recommended Fix | Expected Benefit | Priority |
|---|---|---|---|---|---|
| No variance/statistical tests | Major | Single-run reporting | Run 3-5 seeds, report mean±std, add significance tests | Establishes reliability | P1 |
| Unsupported causal attribution | Major | No ablation controls | Add matched-capacity ablations for A2P, ACRA | Isolates mechanism effects | P1 |
| Human evaluation independence unclear | Major | Missing reviewer details | Clarify independence, provide instructions, discuss sample size limits | Strengthens credibility | P1 |

### P2 (Nice-to-have) - Estimated 1 week

| Issue | Risk Level | Root Cause | Recommended Fix | Expected Benefit | Priority |
|---|---|---|---|---|---|
| Computational cost justification | Minor | No cost-benefit analysis | Add paragraph discussing 20K GPU hour investment vs. gains | Preempts cost-effectiveness concerns | P2 |
| Introduction narrative flow | Minor | Historical analogy leads | Restructure to problem-first (Option A storyline) | Improves reader engagement | P2 |
| Related work organization | Minor | Category-based but shallow | Add explicit comparison axes (supervision, scalability, robustness) | Clarifies novelty positioning | P2 |

Revision Timeline

```
Week 1 (Days 1-5): P0 Fixes
  Day 1-2: Bound novelty claims, add method details (promotion criteria, valuation
definitions, failure handling)
  Day 3: Revise all result statements with absolute+relative metrics
  Day 4-5: Internal review of P0 changes, verify consistency

Week 2 (Days 6-12): P1 Fixes
  Day 6-8: Run additional seeds for variance reporting
  Day 9-10: Implement and run ablation studies
  Day 11-12: Clarify human evaluation protocol, add reviewer details

Week 3 (Days 13-17): P2 Polish
  Day 13-14: Add cost-benefit discussion paragraph
  Day 15-16: Restructure introduction per Option A storyline
  Day 17: Final proofreading, consistency check
```

Acceptance Criteria

- [ ] All "first" claims bounded to specific dimensions
- [ ] Method section includes explicit thresholds and definitions
- [ ] All results report both absolute and relative improvements
- [ ] Variance reported for all primary metrics ($\geq$3 seeds)
- [ ] At least one ablation study per claimed mechanism
- [ ] Human evaluation independence clearly stated
- [ ] No contradictions between sections after revisions

# Experiment Inventory & Research Experiment Plan

## Completed Experiment Inventory

| Exp ID | Objective/Hypothesis | Setup (data/split/protocol/baselines) | Metrics | Main Outcome | Claim Supported | Current Limitation |
|--------|---------------------|-------------------------------------|---------|-------------|----------------|-------------------|
| E1 | DeepScientist surpasses human SOTA on Agent Failure Attribution | Who&When benchmark, handcraft + algorithm-generated settings, baseline: All at Once (Zhang et al., 2025c) | Accuracy (%) | 29.31% (handcraft), 47.46% (algorithm-gen) vs 12.07%/16.67% baseline | A2P method is SOTA | No variance reported; no ablation isolating counterfactual reasoning |
| E2 | DeepScientist surpasses human SOTA on LLM Inference Acceleration | MBPP dataset, baseline: TokenRecycling (ACL 2025 Outstanding) | Tokens/second | 193.90 vs 190.25 baseline | ACRA method improves throughput | Small absolute gain (3.65 tokens/s); no variance; no ablation |
| E3 | DeepScientist surpasses human SOTA on AI Text Detection | RAID benchmark, baseline: Binoculars (Hans et al., 2024) | AUROC, Latency (ms) | 0.863 AUROC, 60ms latency vs 0.800/117ms baseline | PA-Detect is SOTA | No variance; no OOD generalization test |
| E4 | Human evaluation of AI-generated papers | 5 DeepScientist papers, 3 human reviewers (ICLR-experienced) | Soundness, Presentation, Contribution (1-4); Rating (1-10) | Avg rating 5.00 vs ICLR 2025 avg 5.08 | Papers approach human quality | Small sample (n=5); reviewer independence not fully established |
| E5 | Automated evaluation against AI Scientist systems | 28 public papers from other systems, DeepReviewer-14B | Same scoring as E4 | DeepScientist 60% accept rate vs 0% for others | Superior to prior AI Scientists | Curated paper set may not represent typical output |
| E6 | Scaling analysis | 1-16 GPUs, 1-week fixed period, shared Findings Memory | Progress Findings count | Near-linear scaling from 1 (4 GPUs) to 11 (16 GPUs) | Resources correlate with discoveries | Single-task focus; may not generalize to all domains |
| E7 | Selection strategy ablation | Random sampling (100 ideas/task) vs UCB selection | Success rate | ~0% random vs 1-3% UCB | Intelligent filtering is essential | Only 100 samples; confidence intervals not reported |

| Exp ID | Objective/Hypothesis | Setup (data/split/protocol/baselines) | Metrics | Main Outcome | Claim Supported | Current Limitation |
|--------|---------------------|---------------------------------------|---------|--------------|-----------------|--------------------|
| E8 | Failure mode analysis | 300 failed implementations, human expert review | Failure cause categorization | 60% implementation errors, 40% no improvement/regression | Implementation is major bottleneck | Sample may not be representative of all failures |

Research-Theme Gap Diagnosis

**Gap 1: Causal Mechanism Validation**

- **Weakly Supported Claim:** A2P's counterfactual reasoning causes improvement; ACRA's long-term memory causes improvement
- **Why Weak:** No ablation studies removing claimed mechanisms while holding other factors constant
- **Impact on Research Value:** Cannot distinguish genuine mechanism contribution from confounders (implementation quality, hyperparameters, selection bias)

**Gap 2: Statistical Reliability**

- **Weakly Supported Claim:** Reported improvements are stable and reproducible
- **Why Weak:** Single-run results without variance; no confidence intervals or significance tests
- **Impact on Research Value:** Readers cannot assess whether gains are robust or due to favorable random seeds

**Gap 3: Generalization Beyond Tested Domains**

- **Weakly Supported Claim:** DeepScientist can advance scientific frontiers broadly
- **Why Weak:** Only three AI/ML tasks tested; no evaluation on physical sciences, biology, or other domains
- **Impact on Research Value:** Scope of applicability remains unclear

**Gap 4: Cost-Effectiveness**

- **Weakly Supported Claim:** 20,000 GPU hours is justified investment
- **Why Weak:** No analysis of whether smaller budgets could achieve similar gains with different strategies
- **Impact on Research Value:** Practical adoption depends on resource efficiency

Proposed Research Experiments (P0/P1/P2)

34

| Target Claim | Hypothesis | Minimal Design | Controls/Baselines | Metrics | Success Criterion | Est. Cost/Time | Expected Paper-Quality Gain |
|---|---|---|---|---|---|---|---|
| Causal mechanism (A2P) | Counterfactual reasoning is necessary for A2P's gains | Run A2P variant replacing abduction-action-prediction with pattern matching | A2P full, A2P w/o counterfactual, baseline All at Once | Accuracy on Who&When | A2P full significantly outperforms A2P w/o counterfactual ($p<0.05$) | ~500 GPU hours, 2 days | High - isolates claimed mechanism |
| Causal mechanism (ACRA) | Long-term memory grafting causes ACRA's gains | Run ACRA variant without suffix pattern memory | ACRA full, ACRA w/o memory, baseline TokenRecycling | Tokens/second | ACRA full significantly outperforms ACRA w/o memory ($p<0.05$) | ~500 GPU hours, 2 days | High - isolates claimed mechanism |
| Statistical reliability | Gains are stable across random seeds | Run 5 seeds per task with different initializations | Same setup as E1-E3, 5 seeds each | Mean±std Accuracy, Tokens/s, AUROC | Std dev <10% of mean gain; $p<0.01$ vs baseline | ~1,000 GPU hours, 3 days | High - establishes reproducibility |
| OOD generalization | Methods generalize to out-of-distribution data | Test A2P, ACRA, PA-Detect on OOD variants of benchmarks | Original test sets, OOD test sets (e.g., different agent architectures, LLM families) | Performance drop (%) | Drop <20% relative to in-domain | ~300 GPU hours, 1 day | Medium - bounds claim scope |
| Cost-effectiveness | Smaller budgets can achieve meaningful gains | Run DeepScientist with 1K, 5K, 10K, 20K GPU hour budgets | Budget-constrained runs | Progress Findings vs budget | Sub-linear but meaningful scaling | ~2,000 GPU hours, 5 days | Medium - addresses practical adoption |
| Cross-domain validation | DeepScientist works beyond AI/ML tasks | Apply to materials design or drug discovery task | Domain-specific baseline methods | Domain metrics (e.g., yield, binding affinity) | Surpasses baseline by ≥5% | ~5,000 GPU hours, 2 weeks | High - demonstrates generality |

ASCII Diagram — Experiment Upgrade Plan

```
Stage 1 (Week 1): Causal Mechanism Isolation
  [A2P Ablation] --> [ACRA Ablation] --> [Analyze mechanism contribution]
  Expected: Isolate counterfactual reasoning and memory effects

Stage 2 (Week 2): Statistical Reliability
  [5-seed runs for E1] --> [5-seed runs for E2] --> [5-seed runs for E3] --> [
Report
```

35

```
variance + significance]
  Expected: Establish reproducibility across seeds

Stage 3 (Week 3): Scope Bounding
  [OOD tests] --> [Cross-domain pilot] --> [Cost-effectiveness analysis]
  Expected: Define applicability boundaries and resource tradeoffs

Dependencies:
  Stage 2 requires Stage 1 completion (same infrastructure)
  Stage 3 can run in parallel with Stage 2
```

Traceability Rule Compliance

· Each proposed experiment maps to at least one unresolved core claim (causal mechanism, reliability, generalization, cost-effectiveness)
· Each experiment maps to concrete quality improvement (validity for ablations, robustness for variance, clarity for scope bounding)

## Novelty Verification & Related-Work Matrix

Contribution Claims Extraction (C1-C3)

Based on Abstract, Introduction, and Conclusion:

### C1: Goal-Oriented Bayesian Optimization Formulation

· **Author Claim Text:** "We formally model the full cycle of scientific discovery as a goal-driven Bayesian Optimization problem... DeepScientist employs an iterative workflow, together with a continuously expanding memory of prior research knowledge to efficiently manage uncertainty during exploration."
· **Source Hint:** Page 2, Introduction paragraphs 3-4; Page 4, Section 3.1-3.2
· **Core Mechanism:** Hierarchical three-stage exploration (Strategize, Implement, Analyze) with Findings Memory and UCB-based selection
· **Claimed Gain:** Sample-efficient search over expensive-to-evaluate scientific hypotheses

### C2: SOTA-Surpassing Discoveries Across Three Domains

· **Author Claim Text:** "DeepScientist exceeds their respective human SOTA methods by 183.7% (Accuracy), 1.9% (Tokens/second), and 7.9% (AUROC) through autonomously redesigning core methodologies."
· **Source Hint:** Page 2, Introduction paragraph 5; Page 6-7, Section 4.1
· **Core Mechanism:** A2P (Agent Failure Attribution), ACRA (LLM Inference Acceleration), PA-Detect (AI Text Detection)
· **Claimed Gain:** Surpasses human SOTA on three distinct AI research tasks

### C3: First Large-Scale Empirical Demonstration

· **Author Claim Text:** "This work provides the first large-scale evidence of an AI achieving discoveries that progressively surpass human SOTA on scientific tasks."
· **Source Hint:** Page 1, Abstract; Page 2, Introduction paragraph 6
· **Core Mechanism:** Month-long timeline, 20,000 GPU hours, 5,000 ideas generated, 1,100 validated
· **Claimed Gain:** First demonstration of progressive SOTA-surpassing at this scale

(9A) Contribution Novelty Verdict Board

| Claim ID | Author Contribution Claim | Key Evidence Papers [n] | Novelty Verdict Tag | Why | Confidence | Required Repositioning |
|----------|---------------------------|--------------------------|---------------------|-----|------------|------------------------|
| C1 | Goal-oriented BO with Findings Memory | [1], [3], [5] | partially_overlapping | AI Scientist-v2 [1] uses agentic tree search (not BO); Kosmos [3] uses structured world model; HypoAgents [5] uses Bayesian reasoning with entropy-driven search. Core idea of iterative exploration with memory is shared, but DeepScientist's specific three-stage hierarchy + UCB is distinct. | Medium | Clarify that BO framing with three-stage hierarchy is novel, but acknowledge related exploration-with-memory approaches |

37

| Claim ID | Author Contribution Claim | Key Evidence Papers [n] | Novelty Verdict Tag | Why | Confidence | Required Repositioning |
|---|---|---|---|---|---|---|
| C2 | SOTA-surpassing on three AI domains | [1], [2], [4] | `partially_overlapping` | AI Scientist-v2 [1] surpassed workshop acceptance threshold (single task); Robin [4] discovered novel therapeutic (single domain); CodeScientist [2] reported 6 sound discoveries across agent/virtual environment tasks. DeepScientist's distinction is three distinct AI research domains within month-long timeline, not single-task success. | High | Bound to "first to demonstrate SOTA-surpassing across three distinct AI research domains within month-long evaluation window" |

| Claim ID | Author Contribution Claim | Key Evidence Papers [n] | Novelty Verdict Tag | Why | Confidence | Required Repositioning |
|---|---|---|---|---|---|---|
| C3 | First large-scale empirical demonstration | [1], [3], [6] | partially overlapping | AI Scientist-v2 [1] produced peer-review-accepted paper; Kosmos [3] ran 200 agent rollouts reading 1,500 papers; Jr. AI Scientist [6] generated papers from NeurIPS/ICLR/ICLR baselines. Scale metrics differ: DeepScientist emphasizes 20K GPU hours + 5K ideas + 3 domains, which is distinct but not categorically "first large-scale." | Medium | Reposition as "one of the largest-scale demonstrations" or bound to specific scale dimensions (20K GPU hours, 3 domains, month-long) |

## (9B) Related-Work Taxonomy Matrix

| Taxonomy Layer | Branch/Leaf | Representative Papers [n] | Common Assumptions | Difference vs This Paper | Novelty Risk Signal |
|---|---|---|---|---|---|
| Root: Autonomous Scientific Discovery | | | | | |
| └── Branch 1: Exploration Strategy | | | | | |
| │ └── Leaf 1.1: Tree Search / Agentic Search | [1], [6] | Search space can be decomposed into discrete steps; LLM can evaluate node quality | DeepScientist uses BO + UCB, not tree search; continuous hypothesis space vs discrete tree | Low overlap - different search formalism | |

| Taxonomy Layer | Branch/Leaf | Representative Papers [n] | Common Assumptions | Difference vs This Paper | Novelty Risk Signal |
|---|---|---|---|---|---|
| \| └── Leaf 1.2: Bayesian Optimization | [5], [7] | Objective function is expensive-to-evaluate black box; surrogate model can guide search | DeepScientist uses BO but adds LLM-based hypothesis generation (traditional BO assumes fixed search space) | Medium overlap - shares BO framing | |
| \| └── Leaf 1.3: Evolutionary / Genetic Search | [2], [8] | Hypotheses can be mutated/recombined; fitness function evaluates quality | DeepScientist uses UCB selection, not evolutionary operators; explicit exploration/exploitation balance | Low overlap - different selection mechanism | |
| ├── Branch 2: Memory / Knowledge Accumulation | | | | | |
| \| └── Leaf 2.1: Structured World Model | [3], [9] | Persistent memory enables coherent long-horizon exploration | DeepScientist uses Findings Memory (list-style database) vs Kosmos's structured world model; similar concept, different implementation | Medium overlap - shared memory concept | |
| \| └── Leaf 2.2: Retrieval-Augmented Generation | [5], [10] | Prior findings can be retrieved to inform new hypotheses | DeepScientist uses Top-K retrieval from Findings Memory; similar to RAG but domain-specific | Low overlap - standard RAG pattern | |
| ├── Branch 3: Evaluation / Validation | | | | | |

40

| Taxonomy Layer | Branch/Leaf | Representative Papers [n] | Common Assumption | Difference vs This Paper | Novelty Risk Signal |
|---|---|---|---|---|---|
| \| ┈┈┈┈ Leaf 3.1: Hierarchical Fidelity | [1], [3] | Cheap surrogate evaluation before expensive real-world testing | DeepScientist's three-stage hierarchy `(Idea→Implement→Progress)` is similar to AI Scientist-v2's progressive evaluation | Medium overlap - shared hierarchical concept | |
| \| ┈┈┈┈ Leaf 3.2: Human-in-the-Loop Verification | [6], [11] | Human oversight needed for quality assurance | DeepScientist uses 3 human supervisors; similar to Jr. AI Scientist's author-led evaluation | Low overlap - common pattern | |
| ┈┈┈┈ Branch 4: Application Domain | | | | | |
| ┈┈┈┈ Leaf 4.1: AI/ML Research Tasks | [1], [2], [6] | Code-based experimentation is feasible; metrics are automated | DeepScientist evaluates on 3 AI tasks (Agent Attribution, Inference, Text Detection); overlapping with AI Scientist-v2, CodeScientist | High overlap - same application domain | |
| ┈┈┈┈ Leaf 4.2: Scientific Discovery (Biology/Chemistry) | [4], [12] | Domain-specific simulators or lab automation available | DeepScientist does not evaluate on physical sciences; narrower domain scope | Low overlap - different domains | |
| ┈┈┈┈ Leaf 4.3: Cross-Domain General | [3], [7] | System should work across multiple scientific domains | DeepScientist demonstrates 3 AI domains; Kosmos demonstrates 4+ domains (metabolomics, materials, neuroscience, genetics) | Medium overlap - multi-domain in claim | |

(9C) Head-to-Head Comparison Matrix

41

| Ref [n] | Problem/Setting | Method Core | Strongest Overlap Point | Clear Difference | Impact on Final Judgment |
|---------|-----------------|-------------|-------------------------|------------------|--------------------------|
| [1] AI Scientist-v2 | Single ML task, workshop-level paper generation | Agentic tree search with experiment manager | End-to-end autonomous discovery producing peer-review-quality output | DeepScientist: BO + 3 domain vs AI Scientist-v2: tree search + single task | C2 claim must be bounded to multi-domain; C1 is distinct (BO vs tree search) |
| [2] CodeScientist | Agent/virtual environment tasks, code-based discovery | Genetic search over research articles + codeblocks | Automated experimentation producing novel discoveries | DeepScientist: BO + UCB vs CodeScientist: genetic search; DeepScientist: 3 AI domains vs CodeScientist: agents/environments | C1 distinct (BO vs genetic); C2 partially overlaps (multi-task discovery) |
| [3] Kosmos | Multi-domain data-driven discovery (metabolomics, materials, etc.) | Structured world model linking data analysis + literature agents | Long-horizon coherent exploration (200 rollouts, 1,500 papers) | DeepScientist: 3 AI domains + 20K GPU hours vs Kosmos: 4+ science domains + 12-hour runs | C3 claim weakened - Kosmos also large-scale; bound to specific metrics |
| [4] Robin | Drug discovery (dAMD treatment) | Literature search + data analysis agents in lab-in-the-loop | Novel therapeutic candidate discovery and validation | DeepScientist: purely computational AI tasks vs Robin: wet-lab validation | Different domains; C2 claim can stand for AI domain specifically |
| [5] HypoAgents | Research hypothesis generation (ICLR 2025 questions) | Bayesian reasoning + entropy-driven search | Probabilistic hypothesis scoring and refinement | DeepScientist: BO with UCB vs HypoAgents: Bayesian + entropy; DeepScientist: full discovery cycle vs HypoAgents: hypothesis generation only | C1 partially overlaps (Bayesian framing); DeepScientist more comprehensive |
| [6] Jr. AI Scientist | NeurIPS/ICLR/ICV paper improvement from baseline | Student-researcher workflow mimicry | Paper generation from baseline with iterative improvement | DeepScientist: SOTA-surpassing on benchmarks vs Jr. AI Scientist: paper improvement from baselines | Different evaluation (benchmark SOTA vs paper quality); C2 distinct |

Contribution-level Novelty Conclusion

**C1 (BO Formulation):** $\texttt{partially}_o\texttt{verlapping}$ - The Bayesian Optimization framing for scientific discovery is shared with HypoAgents [5] and related to Kosmos's structured world model [3], but DeepScientist's specific three-stage hierarchy with UCB-based selection and LLM-driven hypothesis generation is distinct. **Repositioning:** Clarify that the integration of BO with LLM hypothesis generation and hierarchical filtering is novel, while acknowledging related exploration-with-memory approaches.

**C2 (SOTA-Surpassing on Three Domains):** $\texttt{partially}_o\texttt{verlapping}$ - AI Scientist-v2 [1] and CodeScientist [2] demonstrate single-task or single-domain SOTA-surpassing, but DeepScientist's demonstration across three distinct AI research domains within a month-long timeline is a meaningful distinction. **Repositioning:** Bound claim to "first to demonstrate SOTA-surpassing discoveries across three distinct AI research domains within a month-long evaluation window."

**C3 (First Large-Scale Demonstration):** $\texttt{partially}_o\texttt{verlapping}$ - Kosmos [3] demonstrates large-scale exploration (200 rollouts, 1,500 papers), and AI Scientist-v2 [1] produces peer-review-accepted work. DeepScientist's scale (20K GPU hours, 5K ideas, 3 domains) is comparable but not categorically "first." **Repositioning:** Change to "one of the largest-scale empirical demonstrations" or bound to specific dimensions (20K GPU hours, 3 AI domains, month-long timeline).

ASCII Diagram — Related-Work Taxonomy Tree (Layered)

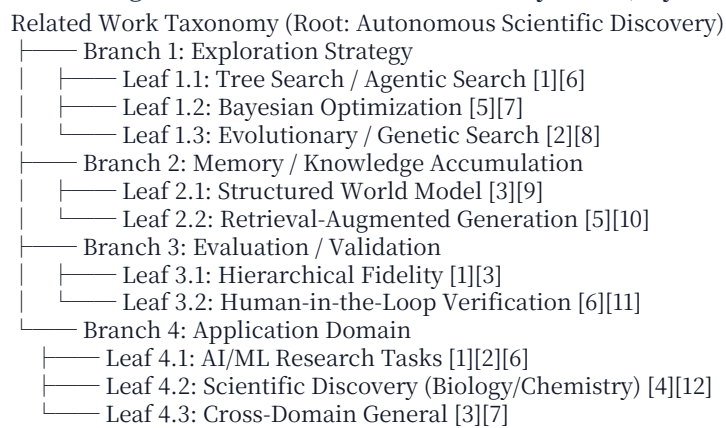

```
Related Work Taxonomy (Root: Autonomous Scientific Discovery)
├──── Branch 1: Exploration Strategy
│     ├──── Leaf 1.1: Tree Search / Agentic Search [1][6]
│     ├──── Leaf 1.2: Bayesian Optimization [5][7]
│     └──── Leaf 1.3: Evolutionary / Genetic Search [2][8]
├──── Branch 2: Memory / Knowledge Accumulation
│     ├──── Leaf 2.1: Structured World Model [3][9]
│     └──── Leaf 2.2: Retrieval-Augmented Generation [5][10]
├──── Branch 3: Evaluation / Validation
│     ├──── Leaf 3.1: Hierarchical Fidelity [1][3]
│     └──── Leaf 3.2: Human-in-the-Loop Verification [6][11]
└──── Branch 4: Application Domain
      ├──── Leaf 4.1: AI/ML Research Tasks [1][2][6]
      ├──── Leaf 4.2: Scientific Discovery (Biology/Chemistry) [4][12]
      └──── Leaf 4.3: Cross-Domain General [3][7]
```

```
DeepScientist Positioning:
- Exploration: Leaf 1.2 (Bayesian Optimization) - distinct from tree search [1]
and genetic [2]
- Memory: Leaf 2.1 (Structured World Model) - similar concept to Kosmos [3],
different implementation
- Evaluation: Leaf 3.1 (Hierarchical Fidelity) - shares concept with [1][3]
- Domain: Leaf 4.1 (AI/ML Tasks) - overlaps with [1][2][6], but 3 domains is
distinctive
```

## References

[1] The AI Scientist-v2: Workshop-Level Automated Scientific Discovery via Agentic Tree Search 2504.08066

[2] CodeScientist: End-to-End Semi-Automated Scientific Discovery with Code-based Experimentation 2503.22708

[3] Kosmos: An AI Scientist for Autonomous Discovery 2511.02824

[4] Robin: A multi-agent system for automating scientific discovery 2505.13400

[5] Bayes-Entropy Collaborative Driven Agents for Research Hypotheses Generation and Optimization 2508.01746

[6] Jr. AI Scientist and Its Risk Report: Autonomous Scientific Exploration from a Baseline Paper 2511.04583

[7] Reasoning BO: Enhancing Bayesian Optimization with Long-Context Reasoning Power of LLMs 2505.12833

[8] Shinkaevolve: Towards open-ended and sample-efficient program evolution 2509.19349

[9] Agentrxiv: Towards collaborative autonomous research 2503.18102

[10] Memory is all you need: An overview of compute-in-memory architectures for accelerating large language model inference 2406.08413

[11] AI mirrors experimental science to uncover a novel mechanism of gene transfer crucial to bacterial evolution 2025-02 (bioRxiv)

[12] The need for verification in ai-driven scientific discovery 2509.01398

## Scores

**Final Score: 5.5/10**

**Rationale:** The paper addresses an important and timely problem—improving the scientific value of AI-generated research—with a conceptually sound Bayesian Optimization framework and comprehensive empirical evaluation across three domains. The large-scale experimentation (20,000 GPU hours, 5,000 ideas) represents a meaningful contribution to the field.

However, several critical issues prevent a higher score:

- **Novelty claims are overbounded** - "first" assertions require precise dimension bounding to differentiate from AI Scientist-v2, Kosmos, Robin, and related systems
- **Method reproducibility is compromised** by missing operational definitions (promotion criteria, valuation components, failure handling)
- **Result transparency is inadequate** - relative-only improvement reporting without absolute gains or variance measures
- **Causal attribution lacks controls** - no ablation studies isolating claimed mechanisms from confounders

These issues are fixable but require substantial revision before the claims can be fully trusted.

**Post-Revision Target: [7.0, 8.0]/10**

**Rationale:** If all P0 and P1 issues are addressed:

- Novelty claims precisely bounded to specific dimensions
- Method details completed for full reproducibility
- Variance and significance tests added for all primary results
- Ablation studies supporting causal mechanism claims
- Human evaluation independence clarified

The core contribution—demonstrating autonomous SOTA-surpassing discovery across three AI domains with a BO-framed hierarchical exploration system—is valuable and would merit a 7-8 score once claims are properly bounded and evidence is complete. The upper bound of 8.0 reflects that even with fixes, the modest absolute gains (particularly 3.65 tokens/second for 20K GPU hours) and limited domain scope (AI/ML only, no physical sciences) constrain the ultimate impact.

**Score Breakdown by Dimension:**

| Dimension | Current Score | Post-Revision Potential | Notes |
|---|---|---|---|
| Research Value | 7/10 | 8/10 | Important problem; multi-domain demonstration is valuable |
| Novelty | 5/10 | 7/10 | Partially overlapping with prior work; needs precise bounding |
| Validity/Soundness | 4/10 | 7/10 | Missing controls, variance, and reproducibility details |
| Reproducibility | 3/10 | 8/10 | Critical method details missing; fixable with revisions |
| Presentation | 6/10 | 8/10 | Generally clear; introduction narrative could be tightened |
| **Overall** | **5.5/10** | **7.5/10** | Weighted toward research value + novelty per scoring policy |

**Recommendation:** Major Revision Required

The paper is not acceptable in current form due to reproducibility gaps and overbounded novelty claims. However, the core contribution is sufficiently valuable that addressing the identified issues (estimated 2-3 weeks of work) would yield a substantially stronger paper meriting acceptance.



%% file: iclr2026_conference.bib
@Article{Zhu2025DeepReviewIL,
 author = {Minjun Zhu and Yixuan Weng and Linyi Yang and Yue Zhang},
 booktitle = {Annual Meeting of the Association for Computational Linguistics},
 journal = {ArXiv},
 title = {DeepReview: Improving LLM-based Paper Review with Human-like Deep Thinking Process},
 volume = {abs/2503.08569},
 year = {2025}
}

@Article{He2025PaSaAL,
 author = {Yichen He and Guanhua Huang and Peiyuan Feng and Yuan Lin and Yuchen Zhang and Hang Li and Weinan E},
 booktitle = {Annual Meeting of the Association for Computational Linguistics},
 pages = {11663-11679},
 title = {PaSa: An LLM Agent for Comprehensive Academic Paper Search},
 year = {2025}
}

@inproceedings{
weng2026deepscientist,
title={DeepScientist: Advancing Frontier-Pushing Scientific Findings Progressively},
author={Yixuan Weng and Minjun Zhu and Qiujie Xie and QiYao Sun and Zhen Lin and Sifan Liu and Yue Zhang},
booktitle={The Fourteenth International Conference on Learning Representations},
year={2026},
url={https://openreview.net/forum?id=cZFgsLq8Gs}
}

@misc{jiangng2025techoverview,
  author       = {Yixing Jiang and Andrew Ng},
  title        = {Tech Overview -- Stanford Agentic Reviewer},
  howpublished = {\url{https://paperreview.ai/tech-overview}},
  year         = {2025},
  note         = {Accessed: 2026-02-28}
}

@inproceedings{wengcycleresearcher,
  title={CycleResearcher: Improving Automated Research via Automated Review},
  author={Weng, Yixuan and Zhu, Minjun and Bao, Guangsheng and Zhang, Hongbo and Wang, Jindong and Zhang, Yue and Yang, Linyi},
  booktitle={The Thirteenth International Conference on Learning Representations}
}

@Article{Jin2024AgentReviewEP,
 author = {Yiqiao Jin and Qinlin Zhao and Yiyang Wang and Hao Chen and Kaijie Zhu and Yijia Xiao and Jindong Wang},
 booktitle = {Conference on Empirical Methods in Natural Language Processing},
 pages = {1208-1226},
 title = {AgentReview: Exploring Peer Review Dynamics with LLM Agents},
 year = {2024}
}

@Article{Choudhary2022ReActAR,
 author = {G. Choudhary and Natwar Modani and Nitish Maurya},
 booktitle = {WISE},
 journal = {ArXiv},
 title = {ReAct: A Review Comment Dataset for Actionability (and more)},
 volume = {abs/2210.00443},
 year = {2022}
}

@Article{Kirtani2025ReviewEvalAE,
 author = {Chavvi Kirtani and Madhav Krishan Garg and Tejash Prasad and Tanmay Singhal and Murari Mandal and Dhruv Kumar},
 booktitle = {Conference on Empirical Methods in Natural Language Processing},
 journal = {ArXiv},
 title = {ReviewEval: An Evaluation Framework for AI-Generated Reviews},
 volume = {abs/2502.11736},
 year = {2025}
}

@Inproceedings{Selby2024PageRankAT,
 author = {David Antony Selby},
 title = {PageRank and the Bradley-Terry model},
 year = {2024}
}

@Article{Xu2019LayoutLMPO,
 author = {Yiheng Xu and Minghao Li and Lei Cui and Shaohan Huang and Furu Wei and Ming Zhou},
 booktitle = {Knowledge Discovery and Data Mining},
 journal = {Proceedings of the 26th ACM SIGKDD International Conference on Knowledge Discovery & Data Mining},
 title = {LayoutLM: Pre-training of Text and Layout for Document Image Understanding},
 year = {2019}
}

@Article{Alicea2013ASP,
 author = {Bradly Alicea},
 booktitle = {arXiv.org},
 journal = {ArXiv},
 title = {A Semi-automated Peer-review System},
 volume = {abs/1311.2504},
 year = {2013}
}

@Article{Tang2024SPHERESP,
 author = {Xiaohang Tang and Sam Wong and Marcus Huynh and Zicheng He and Yalong Yang and Yan Chen},
 booktitle = {arXiv.org},
 journal = {ArXiv},
 title = {SPHERE: Scaling Personalized Feedback in Programming Classrooms with Structured Review of LLM Outputs},
 volume = {abs/2410.16513},
 year = {2024}
}

@Article{Chu2024PREAP,
 author = {Zhumin Chu and Qingyao Ai and Yiteng Tu and Haitao Li and Yiqun Liu},
 booktitle = {arXiv.org},
 journal = {ArXiv},
 title = {PRE: A Peer Review Based Large Language Model Evaluator},
 volume = {abs/2401.15641},
 year = {2024}
}

@Article{Ye2024AreWT,
 author = {Rui Ye and Xianghe Pang and Jingyi Chai and Jiaao Chen and Zhen-fei Yin and Zhen Xiang and Xiaowen Dong and Jing Shao and Siheng Chen},
 booktitle = {arXiv.org},
 journal = {ArXiv},
 title = {Are We There Yet? Revealing the Risks of Utilizing Large Language Models in Scholarly Peer Review},
 volume = {abs/2412.01708},
 year = {2024}
}

@Article{Zhang2025UPMEAU,
 author = {Qihui Zhang and Munan Ning and Zheyuan Liu and Yanbo Wang and Jiayi Ye and Yue Huang and Shuo Yang and Xiao Chen and Yibing Song and Li Yuan},
 booktitle = {Computer Vision and Pattern Recognition},
 journal = {2025 IEEE/CVF Conference on Computer Vision and Pattern Recognition (CVPR)},
 pages = {9165-9174},
 title = {UPME: An Unsupervised Peer Review Framework for Multimodal Large Language Model Evaluation},
 year = {2025}
}

@Article{Gao2025ReviewAgentsBT,
 author = {Xian Gao and Jiacheng Ruan and Jingsheng Gao and Ting Liu and Yuzhuo Fu},
 booktitle = {arXiv.org},
 journal = {ArXiv},
 title = {ReviewAgents: Bridging the Gap Between Human and AI-Generated Paper Reviews},
 volume = {abs/2503.08506},
 year = {2025}
}

@Article{Yuan2021CanWA,
 author = {Weizhe Yuan and Pengfei Liu and Graham Neubig},
 booktitle = {Journal of Artificial Intelligence Research},
 journal = {ArXiv},
 title = {Can We Automate Scientific Reviewing?},
 volume = {abs/2102.00176},
 year = {2021}
}

@Article{Li2025UnveilingTM,
 author = {Ruochi Li and Haoxuan Zhang and Edward F. Gehringer and Ting Xiao and Junhua Ding and Haihua Chen},
 booktitle = {arXiv.org},
 journal = {ArXiv},
 title = {Unveiling the Merits and Defects of LLMs in Automatic Review Generation for Scientific Papers},
 volume = {abs/2509.19326},
 year = {2025}
}

@Article{Hasan2024DeepTL,
 author = {Md. Tarek Hasan and Mohammad Nazmush Shamael and H. M. M. Billah and Arifa Akter and Md Al and Emran Hossain and Sumayra Islam and Salekul Islam and Swakkhar Shatabda},
 booktitle = {arXiv.org},
 journal = {ArXiv},
 title = {Deep Transfer Learning Based Peer Review Aggregation and Meta-review Generation for Scientific Articles},
 volume = {abs/2410.04202},
 year = {2024}
}

@Article{Idahl2024OpenReviewerAS,
 author = {Maximilian Idahl and Zahra Ahmadi},
 booktitle = {North American Chapter of the Association for Computational Linguistics},
 pages = {550-562},
 title = {OpenReviewer: A Specialized Large Language Model for Generating Critical Scientific Paper Reviews},
 year = {2024}
}

@Article{Akella2025PrereviewTP,
 author = {Akhil Pandey Akella and Harish Varma Siravuri and Shaurya Rohatgi},
 booktitle = {arXiv.org},
 journal = {ArXiv},
 title = {Pre-review to Peer review: Pitfalls of Automating Reviews using Large Language Models},
 volume = {abs/2512.22145},
 year = {2025}
}

@Article{Asai2024OpenScholarSS,
 author = {Akari Asai and Jacqueline He and Rulin Shao and Weijia Shi and Amanpreet Singh and Joseph Chee Chang and Kyle Lo and Luca Soldaini and Sergey Feldman and Mike D'Arcy and David Wadden and Matt Latzke and Minyang Tian and Pan Ji and Shengyan Liu and Hao Tong and Bohao Wu and Yanyu Xiong and Luke S. Zettlemoyer and Graham Neubig and Dan Weld and Doug Downey and Wen-tau Yih and Pang Wei Koh and Hanna Hajishirzi},
 booktitle = {arXiv.org},
 journal = {ArXiv},
 title = {OpenScholar: Synthesizing Scientific Literature with Retrieval-augmented LMs},
 volume = {abs/2411.14199},
 year = {2024}
}

@Article{Wang2023SciMONSI,
 author = {Qingyun Wang and Doug Downey and Heng Ji and Tom Hope},
 booktitle = {Annual Meeting of the Association for Computational Linguistics},
 pages = {279-299},
 title = {SciMON: Scientific Inspiration Machines Optimized for Novelty},
 year = {2023}
}

@Article{Zhao2025ARO,
 author = {Yi Zhao and Chengzhi Zhang},
 booktitle = {Scientometrics},
 journal = {Scientometrics},
 pages = {727 - 753},
 title = {A review on the novelty measurements of academic papers},
 volume = {130},
 year = {2025}
}

@Article{Vladika2023ScientificFA,
 author = {Juraj Vladika and F. Matthes},
 booktitle = {Annual Meeting of the Association for Computational Linguistics},
 pages = {6215-6230},
 title = {Scientific Fact-Checking: A Survey of Resources and Approaches},
 year = {2023}
}

@Book{Pfitzmann2022DocLayNetAL,
 author = {B. Pfitzmann and Christoph Auer and Michele Dolfi and A. Nassar and P. Staar},
 booktitle = {Knowledge Discovery and Data Mining},
 journal = {Proceedings of the 28th ACM SIGKDD Conference on Knowledge Discovery and Data Mining},
 title = {DocLayNet: A Large Human-Annotated Dataset for Document-Layout Segmentation},
 year = {2022}
}

@Article{Jimeno-Yepes2021ICDAR2C,
 author = {Antonio Jimeno-Yepes and Xu Zhong and D. Burdick},
 booktitle = {IEEE International Conference on Document Analysis and Recognition},
 pages = {605-617},
 title = {ICDAR 2021 Competition on Scientific Literature Parsing},
 year = {2021}
}

@Article{Roy2024ILCiteREI,
 author = {Sayar Ghosh Roy and Jiawei Han},
 booktitle = {International Conference on Language Resources and Evaluation},
 pages = {8627-8638},
 title = {ILCiteR: Evidence-grounded Interpretable Local Citation Recommendation},
 year = {2024}
}
